\pdfoutput=1

\documentclass[11pt]{article}

\usepackage[]{acl}

\usepackage{csquotes}

\usepackage{hyperref}
\usepackage{array,graphicx}

\usepackage{comment}
\usepackage{amssymb} %
\usepackage{colortbl}
\usepackage{pifont}
\usepackage{amsmath}
\usepackage{tabularx}
\usepackage{multirow}

\usepackage{soul}
\definecolor{lightred}{rgb}{0.98,0.88,0.88}
\definecolor{lightgreen}{rgb}{0.88,0.98,0.88}

\definecolor{lightblue}{rgb}{0.53, 0.81, 0.98}
\definecolor{lightgreen2}{rgb}{0.56, 0.93, 0.56}

\newcommand{\hllightblue}[1]{{\sethlcolor{lightblue}\hl{#1}}}
\newcommand{\hllightgreen}[1]{{\sethlcolor{lightgreen2}\hl{#1}}}

\usepackage{changepage}

\renewenvironment{displayquote}
   {\begin{adjustwidth}{-1em}{-1em}\quote}
   {\endquote\smallskip\end{adjustwidth}}

\newcommand{\greencheck}{\cellcolor{lightgreen}\ding{51}}
\newcommand{\redcross}{\cellcolor{lightred}\ding{55}}

\newcolumntype{Y}{>{\raggedright\arraybackslash}m{\hsize}}

\usepackage{booktabs}

\usepackage{times}
\usepackage{latexsym}

\usepackage[T1]{fontenc}

\usepackage[utf8]{inputenc}

\usepackage[textsize=scriptsize]{todonotes}

\usepackage{inconsolata}

\newcommand{\datasetname}{\mbox{\textsc{X-PARADE}~}}
\newcommand{\datasetphrase}{\textbf{C}ross-lingual \textbf{Par}agraph-level \textbf{A}nalysis of \textbf{D}ivergences and \textbf{E}ntailments}

\newcommand{\isr}[1]{{\color{black}\textbf{#1}}}

\newcommand{\is}[1]{{\hllightgreen{#1}}}

\newcommand{\ns}[1]{{\hllightblue{#1}}}

\newcommand{\samespan}[1]{{#1}}

\newcommand{\finaltokennumber}{106,035~} %

\newcommand{\finalsamplenumberhindi}{191~}
\newcommand{\finalsamplenumberchinese}{199~}
\newcommand{\finalsamplenumber}{576~} %

\newcommand{\FinalNumLanguagePairs}{three~} %
\newcommand{\FinalNumLanguageDirections}{six~} %
\newcommand{\FinalNumLanguages}{four~} %
\newcommand{\FinalNumDirections}{six~} %

\usepackage{microtype}

\usepackage{pdfpages}

\title{X-PARADE: Cross-Lingual Textual Entailment and Information Divergence across Paragraphs}

\author{Juan Diego Rodriguez$^\diamondsuit$ \quad Katrin Erk$^\diamondsuit$$^\spadesuit$ \quad  Greg Durrett$^\diamondsuit$  \\
$^\diamondsuit$ Department of Computer Science \quad $^\spadesuit$ Department of Linguistics \\ The University of Texas at Austin\\
\texttt{juand-r@utexas.edu} \\
}

\begin{document}
\maketitle
\begin{abstract}
Understanding when two pieces of text convey the same information is a goal touching many subproblems in NLP, including textual entailment and fact-checking. This problem becomes more complex when those two pieces of text are in different languages. %
Here, we introduce \datasetname~(\datasetphrase), 
the first cross-lingual dataset of paragraph-level \emph{information divergences}. Annotators label a paragraph in a target language at the span level and evaluate it with respect to a corresponding paragraph in a source language, indicating whether a given piece of information is the same, new, or new but can be inferred. This last notion establishes a link with cross-language NLI. Aligned paragraphs are sourced from Wikipedia pages in different languages, reflecting real information divergences observed in the wild. Armed with our dataset, we investigate a diverse set of approaches for this problem, including token alignment from machine translation, textual entailment methods that localize their decisions, and prompting LLMs. Our results show that these methods vary in their capability to handle inferable information, but they all fall short of human performance.\footnote{Dataset available at \url{https://github.com/juand-r/x-parade}} 

\end{abstract}

\section{Introduction}

The ability to recognize differences in meaning between texts %
underlies many NLP tasks such as natural language inference (NLI), semantic similarity, paraphrase detection, and factuality evaluation. %
Less work exists on the cross-lingual variants of these tasks. However, correctly identifying semantic relations %
between sentences in different languages has a number of useful applications. These include %
estimating the quality of machine translation output \citep{fomicheva2020mlqe}, cross-lingual fact checking \citep{huang-etal-2022-concrete}, and helping Wikipedia editors mitigate discrepancies in content across languages \cite{DBLP:journals/tweb/GottschalkD17}. %
The fact that different languages carve up the world in different ways \citep{saussure1916course,liu-etal-2023-crosslingual} and have different syntactic constraints \citep{keenan1978} may also make these tasks more challenging.

Many of these tasks involve reasoning beyond the sentence level. At the level of paragraphs, it is no longer useful to have coarse labels like ``entailed'' or ``neutral''; instead, we want to capture subtle differences in information content 
\citep{agirre-etal-2016-semeval-2016,briakou-carpuat-2020-detecting, wein-schneider-2021-classifying}. Thus, we focus on the problem of detecting fine-grained span-level \emph{information divergences} between texts across languages. Notably, our notion of information divergences %
differentiates between new information and new information that can be inferred from the source paragraph. 

\begin{figure}[t!]
    \centering
    \includegraphics[width=\linewidth, trim=2mm 100mm 158mm 50mm]{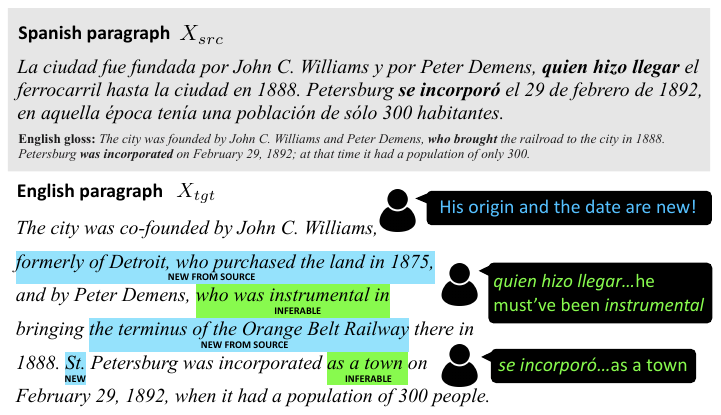}
    \caption{Wikipedia articles written in different languages often contain fine-grained differences in information, such as this paragraph pair taken from the English and Spanish articles on St.~Petersburg, Florida. \datasetname contains fine-grained span-level annotations for content in the target paragraph $X_{\text{tgt}}$ that is \emph{\ns{new}} or \emph{\is{inferable}} given the source paragraph $X_{\text{src}}$.}
    \label{figs:example1}
\end{figure}

\begin{table*}[h]
\centering
\small
\begin{tabular}{lcccccccc}
\toprule
                      & WiCE   &     CLTE-2013 & e-SNLI       &  iSTS        & MS-RTE        &  MLQE-PE      &  REFRESD     & X-PARADE     \\
\midrule
Cross-lingual         &  \redcross     & \greencheck & \redcross    & \redcross    & \redcross     & \greencheck   & \greencheck   & \greencheck   \\
Multiple sentences       &  \greencheck   & \redcross   & \redcross    & \redcross    & \redcross     & \redcross     & \redcross     & \greencheck   \\
Fine-grained annotation &  \greencheck   & \redcross   & \greencheck  & \greencheck  & \greencheck   & \greencheck   & \greencheck   & \greencheck   \\
Entailment relations    &  \greencheck   & \greencheck & \greencheck  & \greencheck  & \greencheck   & \redcross     & \redcross     & \greencheck   \\
\bottomrule
\end{tabular}
\caption{Comparison between \datasetname and related datasets. Ours is the first dataset to provide cross-lingual, paragraph-level annotation of fine-grained entailment.}
\label{tab:datasets-compared}
\end{table*}

This paper presents a dataset called \datasetname: \datasetphrase. Figure~\ref{figs:example1} shows an example English-Spanish paragraph pair, with annotations on how the English paragraph differs from the Spanish paragraph. We see a rich range of inferences being required to understand the target, including effects like \emph{quien hizo llegar} (\emph{who brought}) implying that someone was \emph{instrumental} in bringing. These kinds of subtle cross-lingual divergences are anchored to individual spans in the target paragraph. Finally, unlike prior work that tackled sentence-level comparisons between languages \cite{briakou-carpuat-2020-detecting}, we annotate entire paragraphs. By having larger textual units, we can capture a wider array of divergences and more appropriately model the nuances of cross-sentence context in this task.

We conduct annotation in \FinalNumLanguagePairs language pairs, yielding \FinalNumLanguageDirections directions, using trained annotators from Upwork who went through extensive qualification and feedback rounds. Our dataset is of high quality, with token-level Krippendorff $\alpha$ agreement scores ranging from 0.55 to 0.65, depending on the language pair.

Finally, we benchmark the performance of existing approaches on this problem. No systems in the literature are directly suitable. We compare a diverse set of techniques that solve different aspects of the problem, including token attribution of NLI models, machine translation (MT) alignment and large language models (LLMs). While GPT-4 performs the best, different approaches have different pros and cons and there remains a gap with human performance.

The main contributions of this work are:

\begin{enumerate}
    \item We introduce \datasetname~(\datasetphrase), a dataset for fine-grained cross-lingual divergence detection at the paragraph level, containing \FinalNumLanguages languages and \FinalNumDirections directions (\mbox{\textsc{es-en}}, \mbox{\textsc{en-es}}, \mbox{\textsc{en-hi}}, \mbox{\textsc{hi-en}}, \mbox{\textsc{zh-en}}, \mbox{\textsc{en-zh}}).
    \item We analyze the ability of LLMs and  techniques based on MT alignment and NLI to identify divergences.   %
    We show that the task is non-trivial even for state of the art models. 
\end{enumerate}

\section{Task Setting and Related Work}

\subsection{Task Setting}

Given pairs of paragraphs ($X_{\text{src}}$, $X_{\text{tgt}}$) with some overlapping information, we consider the problem of identifying spans in $X_{\text{tgt}}$ (the target) containing information not present in $X_{\text{src}}$ (the source). $X_{\text{src}}$ and $X_{\text{tgt}}$ are in different languages. %
Our dataset consists of a set of tuples $(X_{\text{src}} , X_{\text{tgt}} , S )$ where $S = \{(t_1, l_1), ... (t_n, l_n)\}$  is a set of labeled spans in the target paragraph $X_{\text{tgt}}$, and $l_i \in Y$ is a label characterizing how $X_{\text{tgt}}$ differs from $X_{\text{src}}$. 
The task is to %
detect both the spans and their label for each $(X_{\text{src}} , X_{\text{tgt}})$. %
Monolingual variants of this task exist, but have mostly concerned themselves with sentence pairs; these include fine-grained textual entailment \citep{brockett2007aligning}, paraphrasing \citep{pavlick-etal-2015-adding}, detection of generation errors \citep{goyal-durrett-2020-evaluating}, including those from LLMs \citep{yue-et-al-2023-automatic-evaluation}, and claim verification \citep{kamoi-etal-2023-wice}.

To determine an appropriate label set $Y$, we reviewed existing taxonomies, 
including taxonomies for paraphrases \citep{vila2014paraphrase} and translations \citep{zhai-etal-2018-construction}. However, these were too fine-grained for our purpose (also including syntactic phenomena), and so we 
use the following mutually-exclusive classes for span-level annotations:\footnote{\label{connotationfootnote}We initially included a fourth category for differences in connotation (e.g., ``slender'' vs ``scrawny''). Given that there were relatively few connotation spans (less than 1\% of tokens), and substantial disagreement between annotators, we decided to remove the connotation labels, and convert them to one of the other three classes, as described in Section ~\ref{sec:adjudication}.}

\begin{enumerate}
    \item \textbf{Same:} The span conveys information nearly identical to some part of the source paragraph. %
    \item \textbf{Inferable:} The span corresponds to a difference in content \emph{inferable from background knowledge or reasoning} given the source paragraph.
    \item \textbf{New:} The span corresponds to a difference in propositional content which cannot be inferred (either new or changed information). %
\end{enumerate}
We did not include a \emph{contradiction} category as in traditional NLI tasks. Explicit contradictions were rare in the naturally-occurring data we observed. However, our taxonomy could be extended to support contradiction for future labeling efforts.

\subsection{Related Tasks}

Here we discuss tasks and datasets which are most closely related to our task. Table~\ref{tab:datasets-compared} compares these datasets and \datasetname along different axes.

\paragraph{Semantic divergence detection} The task of \emph{semantic divergence detection}, i.e., identifying whether cross-lingual text pairs differ in meaning, %
was considered in \citet{vyas-etal-2018-identifying}, but not at the span-level. \citet{wein-schneider-2021-classifying} label semantic divergences between English and Spanish sentences based on their AMR representations, but the distinctions captured are more subtle than what we are aiming for, since some of the subtle distinctions do not affect inference.  %
\citet{briakou-carpuat-2020-detecting} created a dataset, REFRESD, indicating which spans diverge in meaning between English and French sentences sampled from WikiMatrix \citep{schwenk-etal-2021-wikimatrix}. Framed in terms of our taxonomy, %
their dataset involves distinguishing \textbf{same} from \textbf{new or inferable} information; i.e., there is no distinction between information that can be inferred or not.

\paragraph{Textual entailment} Several studies have considered the task of not only predicting entailment relations between sentence pairs, but also detecting which spans contribute to that decision. These tasks differ in terms of the structure and granuality of entailment relations. The MSR RTE dataset \citep{brockett2007aligning} is the RTE-2 data \cite{haim2006second} annotated with span alignment information. The e-SNLI dataset \cite{DBLP:conf/nips/CamburuRLB18} is annotated with spans which explain the relation (entailment, neutral or contradiction) between two sentences. Finally, the Interpretable STS (iSTS) shared task consisted in identifying and aligning spans between two sentences \citep{agirre-etal-2016-semeval-2016} with labels similar to the Natural Logic entailment relations \citep{maccartney-manning-2009-extended}. These studies use monolingual (English) sentences, unlike our work. Of these datasets, only iSTS distinguishes between \textbf{same} and \textbf{inferable} information.

Related to this work is fine-grained and explainable NLI. %
\citet{zaman2022multilingual} use MT alignment to measure the plausibility and faithfulness of token attribution methods for multilingual NLI models. Their work builds on XNLI \citep{conneau-etal-2018-xnli}, which uses translation and is typically handled in a monolingual setting. \citet{stacey2022logical} build sentence-level NLI models by combining span-level predictions with simple rules. Finally, WiCE \citep{kamoi-etal-2023-wice} consists of monolingual document-claim pairs %
with token-level labels for non-supported (i.e., non-entailed) tokens.

There is also a small literature on cross-language %
textual entailment (CLTE), mostly consisting of older techniques \citep{negri-etal-2012-semeval, negri-etal-2013-semeval}. There has been little work following in this vein, and modern neural methods enable us to pursue a more ambitious scope of changes detected.

\paragraph{Other tasks} Two other tasks which also involve finding spans in text pairs are word-level quality estimation for MT, and factuality evaluation of generated summaries \citep{tang-et-al-2023-understanding-factual}. MLQE-PE \citep{fomicheva2020mlqe} and HJQE \citep{yang2022rethink} have been annotated for word-level MT quality estimation. XSumFaith \citep{maynez-etal-2020-faithfulness} 
and CLIFF \citep{cao-wang-2021-cliff} contain annotations of non-factual spans in generated summaries.

\section{Dataset Construction} \label{sec:dataset-construction}

\begin{figure}[t!]
\centering
    \includegraphics[width=\linewidth]{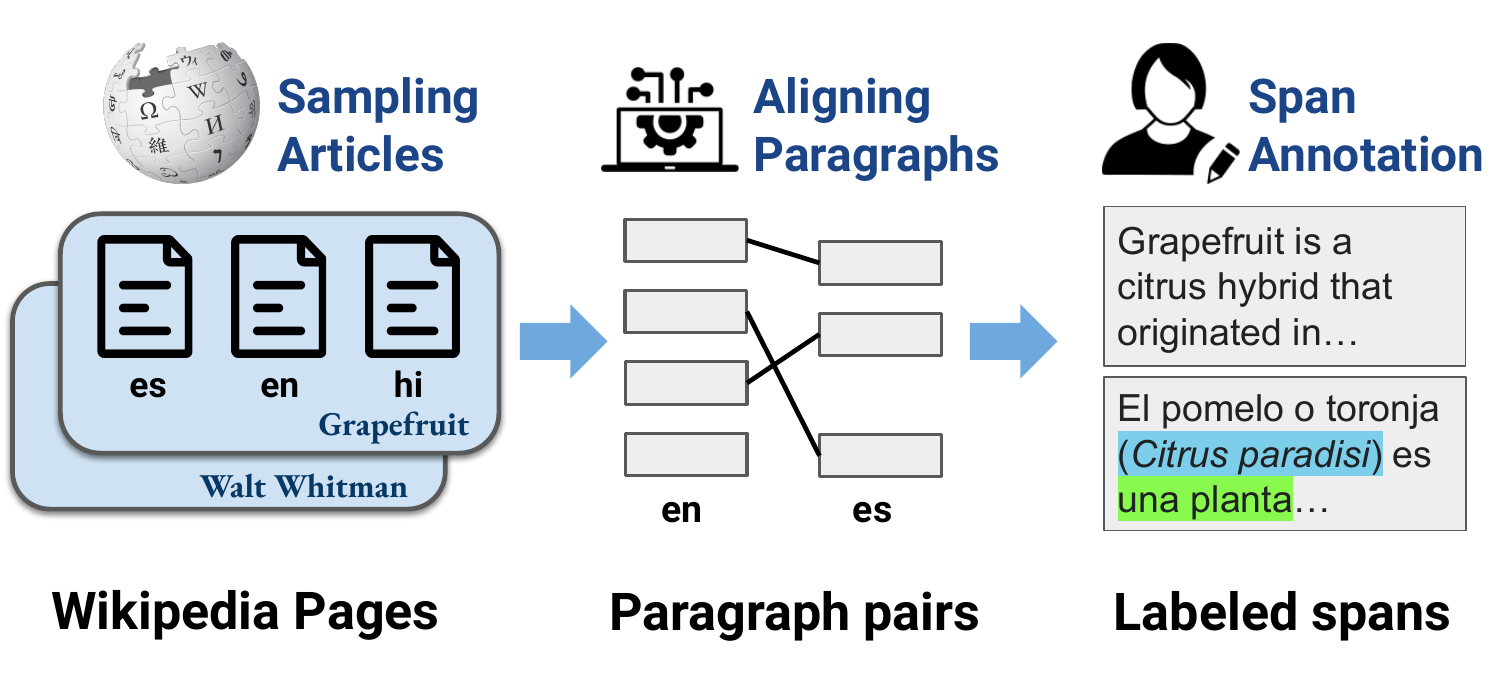}
    \caption{The dataset construction process. Sufficiently similar cross-lingual paragraph pairs are mined from Wikipedia, then annotated by experts. }
    \label{figs:pipeline}
\end{figure}

Our dataset construction pipeline is shown in Figure~\ref{figs:pipeline}. It consists of three stages. We sample from a diverse set of Wikipedia pages, identify paragraph pairs that are sufficiently related but not identical to serve as candidates for our annotation, and present these to annotators to label.  %

\subsection{Data Collection}

\paragraph{Paragraph selection}

Wikipedia pages with versions in English, Spanish, Hindi and Chinese were sampled from the list of pages in CREAK \cite{DBLP:conf/nips/OnoeZCD21} in order to ensure a balanced distribution across topics.  Paragraph alignment between pages was performed by first computing paragraph-paragraph similarities with LaBSE %
\citep{feng-etal-2022-language}, and selecting the set of pairs $\{ (A_i,B_i) \}$ such that $A_i$ and $B_i$ mutually prefer each other over all other paragraphs, ensuring a 1-1 matching.

Finally, one paragraph pair was selected randomly from each article,\footnote{Given the prevalence of summary paragraphs, we re-sampled whenever either of the paragraphs was the first paragraph of the article.} while ensuring similarity scores were distributed uniformly between $0.5$ and $1$.  %
After a manual inspection, we further filtered paragraph pairs by length and similarity score. Additional details are given in Appendix \ref{sec:appendix-dataset-construction}. %

\paragraph{Annotation Process} 

We recruited workers with translation experience between the languages they were annotating. To ensure quality control, workers had to pass a qualification round. %
210 paragraphs were annotated for each language pair in both directions (at an estimated average total time of 84 hours for each language pair). %
The instructions given to annotators are in Appendix \ref{sec:appendix-dataset-instruction} and the annotation interface is shown in Appendix \ref{sec:appendix-dataset-interface}.

Both the adjudicated annotations (described in Section \ref{sec:adjudication}) and each annotator's individual annotations are made publicly available.

\subsection{Inter-annotator Agreement (IAA)} \label{sec:iaa}

Our task involves human judgements about natural language inference, which are known to be subjective \citep{pavlick-kwiatkowski-2019-inherent}. There are many different reasons why annotators may disagree about whether one piece of information entails another \citep{jiang-and-marneffe-2022-investigating}. Here, we evaluate annotator agreement on our task, with a particular focus on the \emph{inferable} category. Some annotators managed to identify a way to infer information in the target while others did not make such inferences and labeled tokens as \emph{new}. In addition some inferences are quite direct, so some annotators labeled them as \emph{same}. For example, there was disagreement over whether ``changes its behavior in spring'' is \emph{new} or \emph{inferable} in the following paragraph pair:%

\begin{displayquote}
\underline{Es:} Las liebres son solitarias...\textbf{Tan solo se producen peleas durante la época de celo} (variable según especies)... Las liebres europeas de sexo masculino apenas comen durante \textbf{este período (primavera)}...\footnote{English gloss: ``Hares are solitary...\textbf{Fights only occur during the mating season} (variable depending on species)... Male European hares hardly eat during \textbf{this period (spring)}...''}
\end{displayquote}

\begin{displayquote}
\underline{En:} Normally a shy animal, the European brown hare \textbf{changes its behavior in spring}...%
\end{displayquote}

In this case, to make the inference that these hares change their behavior in spring, one needs to to link ``este período (primavera)'' (spring) to ``la época de celo'' (mating season), and then realize that hares only fighting during mating season implies a change in their behavior in the spring. %
Additional examples of annotator disagreement over inferable spans are given in Appendix \ref{sec:appendix-inferable-span-disagreement}.

With this context in mind, we compute two measures of inter-annotator agreement. Table~\ref{tab:inter-annotator} shows Krippendorff's $\alpha$\ and token-level macro F1. Krippendorff's $\alpha$ is calculated at the token level following \citet{goyal-etal-2022-snac}. %
Following \citet{briakou-carpuat-2020-detecting} and \citet{deyoung-etal-2020-eraser}, we report the token-level macro F1 score averaged over pairs of annotators (e.g., for three annotators, average over six F1 scores). 

\begin{table}[t!]
\small
    \centering
\begin{tabular}{lcc}
\toprule
& Krippendorff's $\alpha$ & macro F1 \\
\midrule
\mbox{\textsc{en-es}} &   0.657    & 62.5 $\pm$ 2.9   \\
\mbox{\textsc{es-en}} &   0.693    & 63.4 $\pm$ 3.4   \\
\midrule
\mbox{\textsc{en-hi}} &   0.605    & 64.9 $\pm$ 1.4   \\  %
\mbox{\textsc{hi-en}} &   0.570    & 60.4 $\pm$ 1.0   \\  %
\midrule
\mbox{\textsc{en-zh}} &   0.589    & 60.0 $\pm$ 3.4   \\ %
\mbox{\textsc{zh-en}} &   0.637    & 61.3 $\pm$ 4.2   \\ %
\bottomrule
\end{tabular}
\caption{Inter-annotator agreement for X-PARADE. Both Krippendorff's $\alpha$ and macro F1 are calculated at the token level.}
 \label{tab:inter-annotator}
\end{table}

We also examine per-class agreement through sentence-level Krippendorff $\alpha$ scores and through per-class token-level F1 scores averaged over pairs of annotators (Table~\ref{tab:inter-annotator-sentence}). Since we do not have sentence-level annotations, we observe whether each sentence contains a span of a given class or not in order to compute sentence-level Krippendorff $\alpha$ scores for each class. Our annotators strongly agree on content that is \emph{same} or \emph{new}, but have lower agreement about \emph{inferable} annotations. As shown in the example above, this can be attributed to the highly subjective nature of the task of identifying natural language inferences \citep{pavlick-kwiatkowski-2019-inherent,jiang-and-marneffe-2022-investigating}.

\begin{table}[t!]
\small
    \centering
\begin{tabular}{lcc|cc}
\toprule
 \multicolumn{1}{l}{} & \multicolumn{2}{c|}{New} & \multicolumn{2}{c}{Inferable}   \\
  &  $\alpha$ & F1 & $\alpha$ & F1 \\
\midrule
\mbox{\textsc{en-es}} &   0.634 &  84.5 $\pm$ 1.4   & 0.246  & 17.4 $\pm$ 6.9   \\
\mbox{\textsc{es-en}} &   0.664 &  86.9 $\pm$ 1.1   & 0.188  & 17.4 $\pm$ 8.3   \\
\midrule
\mbox{\textsc{en-hi}} &   0.555 & 77.8 $\pm$ 2.4    & 0.253  & 30.2 $\pm$ 2.3   \\ %
\mbox{\textsc{hi-en}} &   0.540 & 75.1 $\pm$ 2.2    & 0.156  & 19.5 $\pm$ 5.4   \\ %
\midrule
\mbox{\textsc{en-zh}} &   0.531 & 79.3 $\pm$ 3.3    & 0.169  & 16.2 $\pm$ 9.0   \\ %
\mbox{\textsc{zh-en}} &   0.572 & 85.7 $\pm$ 3.0    & 0.213  & 14.4 $\pm$ 11.7  \\ %
\bottomrule
\end{tabular}
\caption{Krippendorff $\alpha$ for sentences %
and per-class token-level F1 scores over pairs of annotators.}
 \label{tab:inter-annotator-sentence}
\end{table}

\paragraph{Handling inferable annotations} We observed that annotators were typically precise when they did select inferable tokens (i.e., they had a valid reason for why the token could be inferred). We can therefore take the union of \emph{inferable} tokens annotated by different annotators (with some caveats, discussed in Section~\ref{sec:adjudication}) to arrive at high-precision inferable tokens for our dataset. This results in a natural interpretation for the \emph{inferable} category: \emph{someone} has reason to infer a given span, as exhibited by one of our annotators constructing an inference, which others possibly did not catch. 

Manual inspection of 17 random Spanish-English paragraph pairs where annotators disagreed (given in Appendix \ref{sec:appendix-inferable-span-disagreement}) supports this strategy. Of the 41 \emph{inferable} spans that were disputed, we judged that 29 of them (71\%) were inferable, 5 (12\%) belonged to the \emph{same} class, 4 (10\%) belonged to the \emph{new} class, and 3 (7\%) could have been \emph{inferable} or \emph{new} depending on how much domain-specific background knowledge %
one has in order to judge the span as inferable. Here we accepted a range of inferences as valid, from more direct inferences such as ``\emph{las últimas décadas de la vida}'' $\Rightarrow$``\emph{it is the end of the human life cycle}'', to more indirect inferences such as the example of the European brown hare discussed above.

\subsection{Adjudication} \label{sec:adjudication}

First, we removed any paragraph pairs whenever two annotators rejected the pair as being too dissimilar, or when at least two annotators selected over 95\% of tokens as new. This left 186 paragraph pairs for English-Spanish (11\% removed), \finalsamplenumberhindi paragraph pairs for English-Hindi (9\% removed) and \finalsamplenumberchinese paragraph pairs for English-Chinese (5\% removed). %

We then adjudicate using majority vote at the token level, except when some annotator used the \emph{inferable} label, where we always adjudicate the token as inferable, following the discussion in Section~\ref{sec:iaa}.\footnote{The only exception to this rule is if only one annotator labeled a token as \emph{inferable} while all the others labeled it as \emph{same}; in this case we adjudicate it as \emph{same}, since these are usually near-translations.} %
If \emph{new} and \emph{same} are tied, we break the tie in favor of \emph{new}, with similar logic as to why \emph{inferable} is preferred.
Connotation labels (less than 1\% of the data; see footnote~\ref{connotationfootnote}) are treated as inferable, since manual inspection revealed this class seemed most appropriate for most of them.

\begin{table}[t]
\small
\centering
\begin{tabular}{lcc|cc|cc}
\toprule
      & \multicolumn{2}{c|}{Paragraphs} & \multicolumn{2}{c|}{Sentences} & \multicolumn{2}{c}{Tokens} \\
      & Dev & Test & Dev & Test & Dev & Test \\
\midrule
\mbox{\textsc{en-es}} & 93  & 93 & 343 &  334  & 8565  & 8245 \\
\mbox{\textsc{es-en}} & 93  & 93 & 344 &  304  & 8933  & 8069 \\
\midrule
\mbox{\textsc{en-hi}} & 95  & 96 & 445 & 405  & 11087  &  10413  \\ %
\mbox{\textsc{hi-en}} & 95  & 96 & 388 & 337  & 9560   &  8829  \\ %
\midrule
\mbox{\textsc{en-zh}} & 100  & 99 & 228 & 204  & 7177  & 6938  \\
\mbox{\textsc{zh-en}} & 100  & 99 & 381 & 372  & 9903  & 9638  \\
\bottomrule
\end{tabular}
 \caption{Number of paragraphs, sentences and tokens in the \datasetname dataset. For each pair, both paragraphs were annotated with spans indicating semantic divergence. Each row indicates the number of \{paragraphs, sentences, tokens\} in the target language (e.g., the Spanish language paragraphs, for \mbox{\textsc{en-es}}).}
 \label{tab:dataset_stats}
\end{table}

\subsection{Dataset Statistics}

\datasetname consists of \finalsamplenumber paragraph pairs across \FinalNumLanguagePairs language pairs, with judgments on over \finaltokennumber individual tokens. %
We split the pairs evenly between development and test sets. The number of paragraphs for each language pair are given in Table~\ref{tab:dataset_stats}, and examples of annotated paragraphs can be found in Appendix \ref{sec:appendix-paragraph-examples}.

The distribution of labels over tokens and spans is given in Table~\ref{tab:dataset_stats_labels}.  %

\begin{table}[t]
\small
\centering
\renewcommand{\tabcolsep}{1.4mm}
\begin{tabular}{lccc|ccc}
\toprule
\multicolumn{1}{l}{} & Same & New & Inf & Same & New & Inf \\
\midrule
\multicolumn{1}{l}{} & \multicolumn{3}{c|}{\textbf{\textsc{en-es}}} & \multicolumn{3}{c}{\textbf{\textsc{es-en}}} \\ 
Tokens             & 7797   & 7032   & 1981     & 7351   & 8183  & 1468           \\
Spans              & 791    & 507    & 444      & 776    & 581   & 382           \\
Sentences          & 486    & 464    & 283      & 451    & 477   & 242           \\ 
\midrule
\multicolumn{1}{l}{} & \multicolumn{3}{c|}{\textbf{\textsc{en-hi}}} & \multicolumn{3}{c}{\textbf{\textsc{hi-en}}} \\ 
Tokens         & 9779   & 7034  & 4687       & 9316  & 5858  & 3215          \\  %
Spans          & 702    & 337   & 469        & 678   & 353   & 386           \\  %
Sentences      & 604    & 402   & 394        & 577   & 349   & 315           \\  %
\midrule
\multicolumn{1}{l}{} & \multicolumn{3}{c|}{\textbf{\textsc{en-zh}}} & \multicolumn{3}{c}{\textbf{\textsc{zh-en}}} \\
Tokens         & 6556   & 4851  & 2708     & 6813  & 9378  & 3350            \\
Spans          & 902    & 431   & 733      & 835   & 569   & 687             \\
Sentences      & 351    & 266   & 295      & 409   & 541   & 383             \\ 
\bottomrule
\end{tabular}
 \caption{Distribution of class labels---same (\textbf{Same}), new information (\textbf{New}) and inferable (\textbf{Inf})---over tokens, spans, and sentences in the \emph{target} paragraph for different language pairs in the \mbox{\datasetname}~dataset. \emph{Sentences} indicates the number of sentences containing at least one span in a given class.}
 \label{tab:dataset_stats_labels}
\end{table}

\section{Methods}

While the task of detecting new and inferable information in paragraphs across languages is novel, it relates to ideas from machine translation and textual entailment. Here we describe how to adapt baselines from these areas to assess their performance on this task, as well as prompting LLMs to produce spans (Figure~\ref{figs:methods}). Implementation details can be found in Appendix \ref{sec:appendix-implementation-details}.

\paragraph{Alignment} MT word alignment predicts which words should be aligned across translations; thus words which do not easily align are more likely to present new content not given in the source paragraph. %
By way of approximation, we will assume in these experiments that unaligned tokens fall into the \emph{new} category.

\paragraph{SLR-NLI} SLR-NLI \citep{stacey2022logical} builds on the idea that a \emph{neutral} or \emph{contradiction} relation holds between two sentences only when there is at least one span in the ``hypothesis'' (target) that is not inferable from the premise. Since these spans are exactly the ones containing new information, we use SLR-NLI to predict which spans in the target paragraph are \emph{new}.

\begin{figure}[t]
\centering
    \includegraphics[width=\linewidth,trim=0mm 82mm 197mm 47mm]{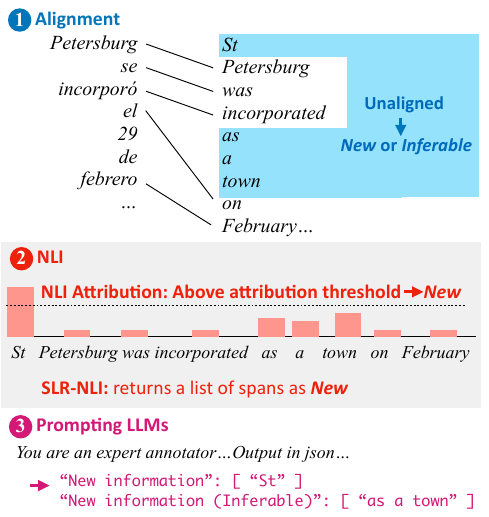}
    \caption{Three of the methods illustrated schematically: (1) the MT-alignment based method attempts to align tokens across texts; tokens which can be aligned are \emph{same}. (2) NLI can be used to either provide attribution scores or spans, identifying tokens which are non-inferable (\emph{new}). (3) LLMs can be prompted to return any desired type of span.}
    \label{figs:methods}
\end{figure}

\paragraph{NLI Attribution} Rather than using the inherently interpretable method of \citet{stacey2022logical}, we can instead use a standard NLI system equipped with a post-hoc interpretation method. We use token attribution methods for NLI models to score the tokens most responsible for a \emph{neutral} classification decision. We compute an attribution score for each token; higher-scoring tokens should be new and not inferable.%

\paragraph{LLMs}

We use one-shot prompting of three state-of-the-art LLMs, \textbf{GPT-3.5-turbo}, \textbf{GPT-4}, and \textbf{Llama-2-chat} \citep{touvron2023llama}, and two explicitly multilingual LLMs, \textbf{BLOOMZ} \citep{muennighoff-etal-2023-crosslingual} and \textbf{XGLM} \citep{lin-etal-2022-shot}. BLOOMZ is an instruction-tuned model, while XGLM is a non-instruction tuned autoregressive LM. We used prompts that specify the annotation task, given in Appendix \ref{sec:appendix-gpt-prompt}. 

The four different methods are compared and summarized in Table~\ref{tab:summary-methods}. Alignment outputs a set of unaligned tokens, while SLR-NLI and NLI token attribution methods produce scores for phrases and tokens, respectively. The LLM generates strings which are then matched to the target paragraph.

\begin{table}[t]
    \centering
    \small
    \begin{tabular}{llll}
    \toprule
    \multicolumn{1}{l}{} & \multicolumn{1}{l}{Outputs} & \multicolumn{1}{l}{Align} & \multicolumn{1}{l}{Translate} \\ 
    \midrule
    Alignment & token set & $\checkmark$  (all) & $\times$ \\
    SLR-NLI & phrase scores & $\checkmark$  (\textsc{en-*}) & $\checkmark$  \\
    NLI Attribution & token scores & $\checkmark$  (\textsc{en-*}) & $\checkmark$  \\ 
    LLM & generated text & $\times$  & $\times$ \\
        \bottomrule
    \end{tabular}
    \caption{Summary of the methods compared. \emph{Align} and \emph{Translate} indicate whether MT alignment and translation are required. Translation is required for the NLI methods since we rely on English-language models.}
    \label{tab:summary-methods}
\end{table}

\section{Results} \label{sec:results}

\begin{table}[t!]
\footnotesize
\renewcommand{\tabcolsep}{1.2mm}
\centering
\begin{tabular}{l|ccc|ccc}
\toprule
    & \multicolumn{3}{c|}{\textbf{ES $\rightarrow$ EN }} & \multicolumn{3}{c}{\textbf{EN $\rightarrow$ ES }} \\
 & P & R & F1 & P & R & F1 \\
\midrule
Majority baseline & 44.6 & 100.0 & 61.7 &  39.8 & 100.0 & 57.0   \\
\midrule
Alignment & 62.3 & 86.1 & 72.3  & 55.4 & 87.4 & 67.8         \\
\midrule
NLI Attr. (IG)    & 64.3 & 78.4   & 70.7  & \emph{51.7} & \emph{80.8} & \emph{63.1}  \\
SLR-NLI           &  67.9 & 78.1  & 72.6  & \emph{60.5} & \emph{64.6} & \emph{62.5}  \\
\midrule
XGLM (7.5B)       & 45.4 & 30.9 & 36.8 & 42.1 & 21.4 & 28.3  \\
Llama-2-chat (7B) & 52.4 & 33.2 & 40.7 & 50.0 & 25.9 & 34.2  \\
GPT-3.5-turbo     & 57.4 & 80.6 & 67.1 & 50.9 & 88.7 & 64.6 \\
GPT-4             & 70.4 & 90.6 & 79.3 & 66.3 & 91.4 & 76.9 \\
\midrule
\multicolumn{7}{c}{w/ Translation to English}\\
\midrule
Llama-2-chat (T)  & 52.3 & 32.3 & 40.0 & \emph{50.8} & \emph{28.5} & \emph{36.5}  \\
GPT-3.5-turbo (T) & 61.0 & 82.1 & 70.0 & \emph{54.7} & \emph{75.2} & \emph{63.3}  \\
GPT-4 (T)         & 72.0 & 89.7 & 79.9 & \emph{63.0} & \emph{80.4} & \emph{70.6}  \\
\hline
\midrule
Human* & 86.8 & 86.5 & \textbf{86.6} & 85.7 & 87.0 & \textbf{86.3} \\
\bottomrule
\end{tabular}
\caption{Precision, recall and F1 scores for new information detection on the English-Spanish test set. Scores in italics indicate methods where both translation and MT alignment was used on the target paragraph.}
\label{tab:results-new-detection-test}
\end{table}

\begin{figure*}[ht!]
\centering
    \includegraphics[width=\linewidth]{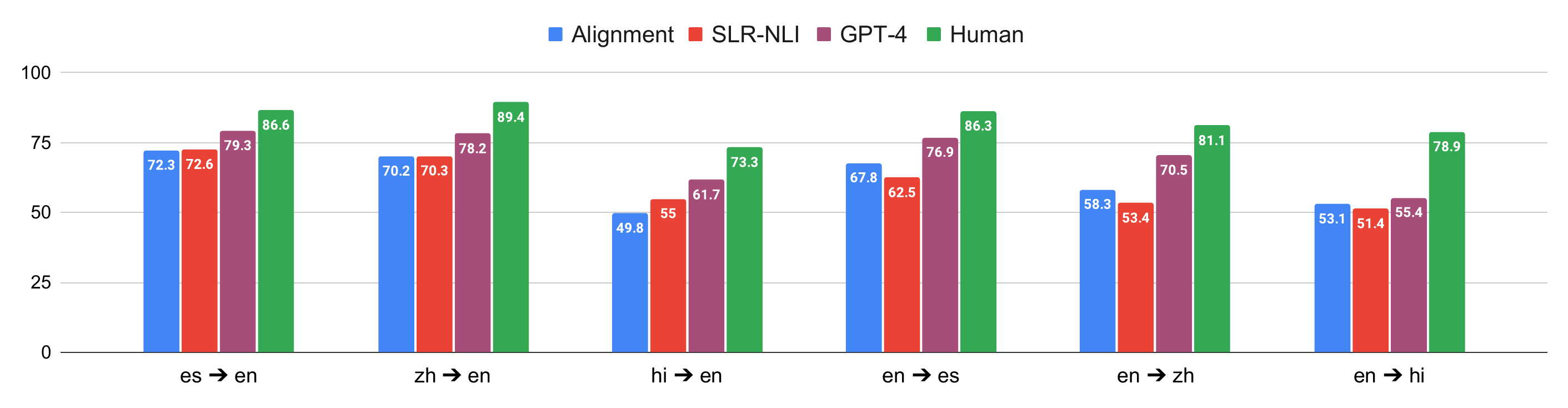}
    \caption{F1 scores for Alignment, SLR-NLI, GPT-4 and human performance on the new information detection task, evaluated on the test set.}
    \label{figs:f1-new-vs-not}
\end{figure*}

\subsection{New information detection (N v. S+I) }
\label{sec:new_info_detection}

Here we discuss results on the binary task of new information detection, i.e., grouping together the classes \emph{same} and \emph{inferable}. Performance on the \mbox{\textsc{en-es}} and \mbox{\textsc{es-en}} test sets are shown in Table~\ref{tab:results-new-detection-test}. We omit scores for BLOOMZ since it substantially underperformed XGLM on every language pair. F1 scores are compared across language pairs in Figure~\ref{figs:f1-new-vs-not}, and full results for the dev set and other language pairs are in Appendix \ref{sec:additional-results}. \textbf{Human*} denotes an estimate of human performance on the task, given by evaluating every annotator against the majority vote of the other annotators, and breaking ties in favor of \emph{new}.

The \mbox{\textsc{en-hi}} and \mbox{\textsc{hi-en}} subsets are harder than \mbox{\textsc{en-es}} and \mbox{\textsc{es-en}}; one possible explanation for this is %
the relative scarcity of Hindi web text, which affects all the NLP components we use (alignment, translation, language models). For every language pair, GPT-4 achieved the highest F1-scores, but there is still a gap in performance compared to humans.\footnote{Since GPT-4 is a closed, proprietary model, we believe there is substantial room to improve performance on this benchmark from the perspective of open models.} %
GPT-3.5-turbo struggles at the task, with scores similar to or worse than the non-LLM methods. XGLM (7.5B) and Llama-2-chat (7B) do worse than the majority-vote baseline. %
This is due to poor instruction-following capacity: we found they often copy from both paragraphs, and sometimes translate them. Both behaviors result in spans that cannot be matched with text in the target. Alignment is surprisingly effective, performing similarly to SLR-NLI for \mbox{\textsc{es-en}}. On the other hand, for \textsc{hi-en}, SLR-NLI outperforms Alignment by 5 points.

\paragraph{Does translating into English improve LLM performance?} When the source language was Spanish (\mbox{\textsc{es-en}}), we observed a small improvement when giving GPT-3.5-turbo translations of the source paragraph (67.1 to 70.0); for \mbox{\textsc{hi-en}} the improvement was more substantial (43.4 to 53.0), and for \mbox{\textsc{zh-en}} using translations had almost no effect. For the \mbox{\textsc{en-*}} language pairs, translating the target paragraph to English did not help GPT-3.5-turbo in most cases; this is likely due to errors in mapping the tokens back to the target language with the MT aligner.
Translating to English did not help GPT-4, which already seems to have strong multilingual capabilities, or Llama-2-chat, which struggled to follow instructions regardless of the language.

\subsection{Inferable Spans}

Both Alignment and NLI Attribution methods only return binary predictions, and so we cannot use them to distinguish inferable spans from \emph{new} or \emph{same}. %
Where do inferable spans fall?  Intuitively, the perfect NLI classifier should fail to distinguish between \emph{same} and \emph{inferable} (predicting the negative class for both) since both lead to entailment. Alignment, on the other hand, should group \emph{new} and \emph{inferable} together in the positive class, since only tokens which are near-perfect translations of each other should align.

Unfortunately, this straightforward picture is not reflected in the system behavior. For the \mbox{\textsc{es-en}} dev set, we noticed that Alignment predicts the positive class for 80.6\% of inferable tokens. However, NLI Attribution predicts the negative class for only 33.4\% of inferable tokens. If both Alignment and NLI Attribution were working perfectly according to our intuitions we would expect all Alignment predictions to be Positive, and all of the NLI attributions to be Negative. %

\subsection{Three-way Divergence Classification with GPT-4}

\begin{table}[t!]
    \small
    \centering
    \setlength{\tabcolsep}{4pt}
    \begin{tabular}{@{}lccc|c@{}}
    \toprule
          & Same & New & Inf & Total \\
    \midrule
    Same  & 2941 & 498  & 241 & 3680 \\
    New   & 405  & 3087 & 108 & 3600 \\
    Inf   & 153  & 515  & 121 & 789 \\
    \midrule
    Total & 3499 & 4100 & 470 \\
    \bottomrule
    \end{tabular}
\caption{Confusion matrix for GPT-4's predictions on the three-way task, on the \mbox{\textsc{es-en}} test set. Rows are the true class labels and columns are predicted labels.} 
    \label{tab:confusion-matrix}
\end{table}

Here we present results on the full divergence taxonomy by prompting GPT-4 with one example including both \emph{inferable} and \emph{new} spans.

We first analyze the raw predictions made by GPT-4 (Table~\ref{tab:confusion-matrix}). %
We note that GPT-4 predicts the inferable label far less frequently than its frequency in our dataset (470 vs 789), and that many predictions are actually same (50\%) or new (23\%). However, it is able to follow the task format and achieves strong performance on \emph{same} and \emph{new} tokens, as suggested by our results in Section~\ref{sec:new_info_detection}.

One example of an incorrectly assigned \emph{inferable} label, is shown below, with GPT-4's prediction highlighted in green:

\begin{displayquote}
\underline{Es:} Los elementos geológicos de Fobos se han nombrado en memoria de astrónomos relacionados con el satélite\footnote{Gloss: ``The geological features of Phobos have been named in memory of astronomers associated with the satellite''}
\end{displayquote}

\begin{displayquote}
\underline{En:} \is{Geological features on Phobos are named after astronomers who studied Phobos} %
\end{displayquote}

\noindent ``Geological...astronomers'' should have been labeled \emph{same} as it closely matches the Spanish. %

\paragraph{Comparison to human performance} Next, we compare GPT-4 against human performance (Human*), which is estimated similarly to Section~\ref{sec:new_info_detection} (except that since it was three-way classification we used the same adjudication procedure as in Section~\ref{sec:adjudication}). For the three-way task, overall performance is slightly lower than human performance (Table~\ref{tab:gpt-4-vs-human}) for all language pairs. %

\begin{table}[t!]
\small
\renewcommand{\tabcolsep}{1.2mm}
\centering
\begin{tabular}{l|l|lll}
\toprule
             &             & P  & R & F1 \\
\midrule
\multirow{2}{*}{\textsc{en-es}}       & GPT-4       & 62.1 &	59.5 &	58.9 \\
            & Human*      & $69.2_{\tiny{\pm 3.5}}$  &  $64.7_{\pm \tiny{2.8}}$ & $64.6_{\pm \tiny{2.4}}$ \\

\multirow{2}{*}{\textsc{es-en}}       & GPT-4       & 61.7 &	60.3 &	60.4 \\
            & Human*      & $70.3_{\tiny{\pm 3.7}}$  &  $65.3_{\pm \tiny{3.2}}$ & $65.1_{\pm \tiny{2.8}}$ \\
\midrule

\multirow{2}{*}{\textsc{en-hi}}       & GPT-4       & 51.3  &  52.4  &  49.4 \\ %
            & Human*      & $66.2_{\tiny{\pm 0.6}}$  & $65.8_{\tiny{\pm 1.2}}$ & $65.6_{\tiny{\pm 1.0}}$ \\ %

\multirow{2}{*}{\textsc{hi-en}}       & GPT-4       &  52.8 &  55.2.	 &  50.6 \\ %
            & Human*      & $61.8_{\tiny{\pm 0.9}}$  & $61.5_{\tiny{\pm 0.3}}$ & $61.3_{\tiny{\pm 0.8}}$ \\ %
\midrule

\multirow{2}{*}{\textsc{en-zh}}       & GPT-4       & 51.8  &  54.0  &  51.4 \\
            & Human*      & $62.6_{\tiny{\pm 0.9}}$  & $61.1_{\tiny{\pm 1.0}}$ & $59.5_{\tiny{\pm 2.4}}$ \\ %

\multirow{2}{*}{\textsc{zh-en}}       & GPT-4       &  57.7 &  56.0	 &  55.4 \\
            & Human*      & $67.3_{\tiny{\pm 2.0}}$  & $65.0_{\tiny{\pm 3.1}}$ & $62.8_{\tiny{\pm 3.2}}$ \\ %
\bottomrule
\end{tabular}
\caption{GPT-4 vs estimated human performance on the three-way classification task; the scores are macro precision, recall and F1 scores on the test set.}
\label{tab:gpt-4-vs-human}
\end{table}

\begin{table}[t!]
\small
\renewcommand{\tabcolsep}{1.2mm}
\centering
\begin{tabular}{l|l|lll}
\toprule
             &             & P  & R & F1 \\
\midrule
\multirow{2}{*}{\textsc{en-es}}       & GPT-4       & 33.2 & 13.1 & 18.8  \\
            & Human*      & $41.2_{\tiny{\pm 13.9}}$  &  $22.9_{\pm \tiny{10.6}}$ & $25.7_{\pm \tiny{5.5}}$ \\

\multirow{2}{*}{\textsc{es-en}}       & GPT-4       & 25.7 & 15.3 & 19.2  \\
            & Human*      & $42.0_{\tiny{\pm 15.3}}$  &  $22.7_{\pm \tiny{11.6}}$ & $24.9_{\pm \tiny{6.9}}$ \\
\midrule

\multirow{2}{*}{\textsc{en-hi}}       & GPT-4       & 29.1  & 8.9 & 13.7 \\ %
            & Human*      & $33.0_{\tiny{\pm 3.3}}$  & $33.1_{\tiny{\pm 7.3}}$ & $32.2_{\tiny{\pm 2.7}}$ \\ %

\multirow{2}{*}{\textsc{hi-en}}       & GPT-4       & 23.4  & 9.5 & 13.5  \\ %
            & Human*      & $19.4_{\tiny{\pm 0.6}}$  & $19.8_{\tiny{\pm 7.3}}$ & $18.9_{\tiny{\pm 4.3}}$ \\ %
\midrule

\multirow{2}{*}{\textsc{en-zh}}       & GPT-4       & 19.9  &  8.0  &  11.4 \\
            & Human*      & $28.8_{\tiny{\pm 6.4}}$  & $20.1_{\tiny{\pm 13.9}}$ & $18.9_{\tiny{\pm 9.2}}$  \\ %

\multirow{2}{*}{\textsc{zh-en}}       & GPT-4       &  30.3 &  13.3	 &  18.5 \\
            & Human*      & $36.2_{\tiny{\pm 10.7}}$  & $21.4_{\tiny{\pm 17.4}}$ & $19.5_{\tiny{\pm 10.5}}$ \\ %
\bottomrule
\end{tabular}
\caption{Performance on \emph{inferable} tokens (GPT-4 vs estimated human performance) on the test set.}
\label{tab:gpt-4-vs-human-inferable}
\end{table}

Finally, Table~\ref{tab:gpt-4-vs-human-inferable} compares GPT-4 and estimated human performance at classifying \emph{inferable} tokens. GPT-4 performs worse than Human*, mainly due to low recall. %
Due to the subjectivity mentioned earlier in Section~\ref{sec:iaa}, it is difficult to obtain an accurate measure of human performance. However, given our analysis of the adjudicated results in Section~\ref{sec:iaa} and Appendix \ref{sec:appendix-inferable-span-disagreement}, we believe that achieving high \emph{precision} of inferable tokens should be possible, even if recall is low, and GPT-4 is far below human performance at this aspect of the task.

\section{Conclusion}

We present \datasetname, a new dataset of cross-lingual paragraph pairs (English-Spanish, English-Hindi, English-Chinese), annotated for semantic divergences at the span-level. Although the task features subjectivity, the analysis of our annotation shows that %
decisions by the annotators were well-justified. We show that while some of these fine-grained differences can be detected by GPT-4, there is still a gap with human performance. We believe that this dataset can be useful for benchmarking the inferential capabilities of multilingual LLMs and analyzing how textual entailment systems can identify information divergences cross-lingually.

\section*{Limitations}

We only compared languages from two different language families (Indo-European and Sino-Tibetan); future work could surface different kinds of differences, reflective either of cultural or typological differences (for an example in Malagasy, see \citet{keenan1978}). Our focus was also on locating inferable or new information, but further work could expand on this to include other aspects such as structuring of information (e.g., discourse markers) and whether information is contradictory rather than merely new. Further, we noted that inferences annotated in \datasetname are sometimes subjective and can take many different forms. Future work could try to further understand the kinds of inferences being made, building on prior work such as \citet{joshi-etal-2020-taxinli} and \citet{jiang-and-marneffe-2022-investigating}.

We explored several baselines for the task, but the methods (e.g., Alignment, NLI Attribution) were not well-suited to distinguish \emph{inferable} from \emph{new} or \emph{same} spans. We hope to see the development of new methods designed explicitly for this task; we believe that better trained cross-lingual NLI systems could potentially be effective here.

Finally, future work could seek to understand why LLMs classify spans as \emph{inferable}. To what extent is it drawing from its parametric knowledge? Given that GPT-4 has seen all of Wikipedia, what constitutes ``background knowledge'' for LLMs and for people is very different. Future work could consider forcing GPT-4 to explain itself (as in chain-of-thought prompting), or explore different structures for how it should generate the data (e.g., forcing it to generate the text spans relevant to the inference).

\section*{Acknowledgments}
Thanks to anonymous reviewers for their helpful feedback. Thanks to the Upwork workers who conducted our annotation task: Isabel Botero, Priya Dabak, Rohan Deshmukh, Fan Feng, Priyanka Ganage, Lin Hongxinnn, Ailin Larossa, John Payne, Tan Wang, Ashish Yadav, and others. This work was partially supported by NSF CAREER Award IIS-2145280, by a gift from Amazon, and by Good Systems,\footnote{\url{https://goodsystems.utexas.edu/}} a UT Austin Grand Challenge to develop responsible AI technologies.

\bibliography{anthology,custom}
\bibliographystyle{acl_natbib}

\newpage
\appendix

\section{Additional Results} \label{sec:additional-results}

Additional results on new information detection are given below in Tables~\ref{tab:results-new-detection-dev-en-es}, \ref{tab:results-new-detection-dev-hi}, and \ref{tab:results-divergence-detection-test-hi}.

\begin{table}[h]
\footnotesize
\renewcommand{\tabcolsep}{1.2mm}
\centering
\begin{tabular}{l|ccc|ccc}
\toprule
    & \multicolumn{3}{c|}{\textbf{ES $\rightarrow$ EN }} & \multicolumn{3}{c}{\textbf{EN $\rightarrow$ ES }} \\
 & P & R & F1 & P & R & F1 \\
\midrule
Majority baseline & 51.3 & 100.0 & 67.8 &  43.7 & 100.0 & 60.9    \\
\midrule
Alignment & 67.4 & 87.6 & 76.1 &  58.2  & 87.7 & 70.0  \\
\midrule
NLI Attr. (IG)  & 66.2 & 78.0 & 71.6  & \emph{53.9} & \emph{80.9} & \emph{64.7}  \\
SLR-NLI         & 69.3 & 76.6 & 72.8  & \emph{62.9} & \emph{70.6} & \emph{66.5}  \\
\midrule
XGLM (7.5B) & 58.3 & 33.8 & 42.8 & 44.5 & 18.5 & 26.1 \\
Llama-2-chat (7B) & 57.5 & 33.5 & 42.4 & 49.0 & 24.0 & 32.2 \\
GPT-3.5-turbo  & 63.6 & 77.7 & 69.9 & 53.7 & 82.9 & 65.2 \\
GPT-4          & 75.3 & 91.4 & 82.6 & 72.2 & 89.4 & 79.9 \\
\midrule
\multicolumn{7}{c}{w/ Translation to English}\\
\midrule
Llama-2-chat (T)  & 63.4 & 34.1 & 44.3 & \emph{55.4} & \emph{28.0} & \emph{37.2}  \\
GPT-3.5-turbo (T) & 63.8 & 82.3 & 71.9 & \emph{56.2} & \emph{73.7} & \emph{63.7}  \\
GPT-4 (T)         & 73.8 & 89.2 & 80.7 & \emph{66.6} & \emph{79.3} & \emph{72.4}  \\
\hline
\midrule
Human* & 90.5 & 90.8 & \textbf{90.7} & 87.1  &  88.1 & \textbf{87.6} \\
\bottomrule
\end{tabular}
\caption{Precision, recall and F1 scores for new information detection on the English-Spanish dev set.}
\label{tab:results-new-detection-dev-en-es}
\end{table}

\begin{table}[h]
\footnotesize
\renewcommand{\tabcolsep}{1.2mm}
\centering
\begin{tabular}{l|ccc|ccc}
\toprule
    & \multicolumn{3}{c|}{\textbf{HI $\rightarrow$ EN }} & \multicolumn{3}{c}{\textbf{EN $\rightarrow$ HI }} \\
 & P & R & F1 & P & R & F1 \\
\midrule
Majority baseline & 35.9 & 100.0 & 52.9 &  34.9 & 100.0 & 51.7   \\ %
\midrule
Alignment         & 45.9 & 82.5  & 59.0 &  42.7 & 85.0 & 56.8   \\
\midrule
NLI Attr. (IG)    & 49.1 & 88.4 & 63.1  & \emph{45.7} & \emph{61.5} & \emph{52.4}  \\
SLR-NLI           & 58.4 & 79.8 & 67.5  & \emph{50.4} & \emph{57.9} & \emph{53.9}  \\
\midrule
XGLM (7.5B)       & 31.5 & 20.9 & 25.1 & 41.2 & 24.2 & 30.5 \\
Llama-2-chat (7B) & 37.2 & 39.1 & 38.1 & 38.2 & 33.1 & 35.5 \\
GPT-3.5-turbo     & 42.1 & 72.2 & 53.2 & 41.2 & 64.7 & 50.4 \\
GPT-4             & 57.7 & 95.6 & 69.6 & 52.4 & 68.9 & 59.5 \\
\midrule
\multicolumn{7}{c}{w/ Translation to English}\\
\midrule
Llama-2-chat (T)  & 46.8 & 36.2 & 40.8 & \emph{45.1} & \emph{26.4} & \emph{33.3} \\
GPT-3.5-turbo (T) & 51.4 & 83.8 & 63.7 & \emph{45.3} & \emph{58.3} & \emph{51.0} \\
GPT-4 (T)         & 56.0 & 94.4 & 70.3 & \emph{50.9} & \emph{61.5} & \emph{55.7} \\
\hline
\midrule
Human*            & 64.5 & 85.1 & \textbf{72.6} & 66.3 & 84.4 & \textbf{73.5}   \\ %
\bottomrule
\end{tabular}
\caption{Precision, recall and F1 scores for new information detection on the English-Hindi dev set.}
\label{tab:results-new-detection-dev-hi}
\end{table}

\begin{table}[h]
\footnotesize
\renewcommand{\tabcolsep}{1.2mm}
\centering
\begin{tabular}{l|ccc|ccc}
\toprule
    & \multicolumn{3}{c|}{\textbf{HI $\rightarrow$ EN }} & \multicolumn{3}{c}{\textbf{EN $\rightarrow$ HI }} \\
 & P & R & F1 & P & R & F1 \\
\midrule
Majority baseline & 27.4 & 100.0 & 43.1  &  30.4 & 100.0 & 46.6   \\ %
\midrule
Alignment         & 35.8 & 81.8  & 49.8 &  38.3 & 86.5  &  53.1  \\
\midrule
NLI Attr. (IG)    & 37.8  & 88.5  & 52.9  & \emph{43.0} & \emph{64.9} & \emph{51.8}  \\
SLR-NLI           & 46.2  & 67.9  & 55.0  & \emph{46.3} & \emph{57.8} & \emph{51.4}  \\
\midrule
XGLM (7.5B)       & 27.7 & 24.7 & 26.1 & 43.1 & 29.0 & 34.7 \\
Llama-2-chat (7B) & 28.8 & 37.7 & 32.6 & 31.9 & 32.2 & 32.1 \\
GPT-3.5-turbo     & 31.6 & 69.5 & 43.4 & 33.6 & 62.5 & 43.7 \\
GPT-4             & 46.4 & 92.1 & 61.7 & 47.4 & 66.7 & 55.4 \\
\midrule
\multicolumn{7}{c}{w/ Translation to English}\\
\midrule
Llama-2-chat (T)  & 34.1 & 31.9 & 33.0 & \emph{39.2} & \emph{24.6} & \emph{30.2} \\
GPT-3.5-turbo (T) & 38.7 & 84.0 & 53.0 & \emph{40.4} & \emph{55.7} & \emph{46.9} \\
GPT-4 (T)         & 45.9 & 91.5 & 61.1 & \emph{50.3} & \emph{62.5} & \emph{55.7} \\
\hline
\midrule
Human* & 65.6 & 85.8 & \textbf{73.9} & 66.5 & 86.4 & \textbf{74.8}  \\ %
\bottomrule
\end{tabular}
\caption{Precision, recall and F1 scores for new information detection on the English-Hindi test set.}
\label{tab:results-divergence-detection-test-hi}
\end{table}

\begin{table}[h]
\footnotesize
\renewcommand{\tabcolsep}{1.2mm}
\centering
\begin{tabular}{l|ccc|ccc}
\toprule
    & \multicolumn{3}{c|}{\textbf{ZH $\rightarrow$ EN }} & \multicolumn{3}{c}{\textbf{EN $\rightarrow$ ZH }} \\
 & P & R & F1 & P & R & F1 \\
\midrule
Majority baseline & 47.4 & 100.0 & 64.3 &  33.2 & 100.0 & 49.8   \\ %
\midrule
Alignment         & 58.1 & 86.2  & 69.4 &  44.2 & 81.8  & 57.4   \\
\midrule
NLI Attr. (IG)    & 57.9 & 91.1 & 70.8  & \emph{43.0} & \emph{71.8} & \emph{53.8}  \\
SLR-NLI           & 65.4 & 79.0 & 71.5  & \emph{53.1} & \emph{53.7} & \emph{53.3}  \\
\midrule
XGLM (7.5B)       & 49.8 & 34.8 & 41.0 & 33.7 & 33.6 & 33.7 \\
Llama-2-chat (7B) & 52.3 & 34.5 & 41.6 & 33.7 & 38.5 & 36.0 \\
GPT-3.5-turbo     & 58.6 & 76.8 & 66.4 & 39.5 & 77.0 & 52.2 \\
GPT-4             & 69.4 & 90.0 & 78.3 & 56.0 & 89.3 & 68.8 \\
\midrule
\multicolumn{7}{c}{w/ Translation to English}\\
\midrule
Llama-2-chat (T)  & 56.8 & 34.8 & 43.1 & \emph{39.0} & \emph{25.9} & \emph{31.1} \\
GPT-3.5-turbo (T) & 55.2 & 81.4 & 65.8 & \emph{41.3} & \emph{68.3} & \emph{51.5} \\
GPT-4 (T)         & 63.6 & 88.1 & 73.9 & \emph{47.8} & \emph{72.0} & \emph{57.4} \\
\hline
\midrule
Human*            & 89.5 & 88.1 & \textbf{88.6} & 83.0 & 83.9 & \textbf{82.9} \\
\bottomrule
\end{tabular}
\caption{Precision, recall and F1 scores for new information detection on the English-Chinese dev set.}
\label{tab:results-new-detection-dev-zh}
\end{table}

\begin{table}[h]
\footnotesize
\renewcommand{\tabcolsep}{1.2mm}
\centering
\begin{tabular}{l|ccc|ccc}
\toprule
    & \multicolumn{3}{c|}{\textbf{ZH $\rightarrow$ EN }} & \multicolumn{3}{c}{\textbf{EN $\rightarrow$ ZH }} \\
 & P & R & F1 & P & R & F1 \\
\midrule
Majority baseline & 48.6 & 100.0  & 65.4 &  35.6 & 100.0  & 52.5   \\ %
\midrule
Alignment         & 59.1 & 86.5  & 70.2 &  45.9 & 79.9  & 58.3   \\
\midrule
NLI Attr. (IG)    & 58.4 & 91.6 & 71.3  & \emph{44.5} & \emph{73.6} & \emph{55.5}  \\
SLR-NLI           & 66.2 & 75.0 & 70.3  & \emph{54.6} & \emph{51.5} & \emph{53.0}  \\
\midrule
XGLM (7.5B)       & 51.1 & 35.1 & 41.6 & 39.1 & 35.2 & 37.0 \\
Llama-2-chat (7B) & 51.9 & 36.7 & 43.0 & 36.8 & 40.6 & 38.7 \\
GPT-3.5-turbo     & 58.3 & 80.8 & 67.7 & 40.6 & 73.0 & 52.2 \\
GPT-4             & 68.5 & 91.1 & 78.2 & 58.3 & 88.3 & 70.5 \\
\midrule
\multicolumn{7}{c}{w/ Translation to English}\\
\midrule
Llama-2-chat (T)  & 57.1 & 34.3 & 42.9 & \emph{43.0} & \emph{31.2} & \emph{36.2} \\
GPT-3.5-turbo (T) & 58.4 & 82.6 & 68.4 & \emph{45.4} & \emph{70.5} & \emph{55.2} \\
GPT-4 (T)         & 65.5 & 91.4 & 76.3 & \emph{53.9} & \emph{73.2} & \emph{62.1} \\
\hline
\midrule
Human*            & 90.7 & 88.8 & \textbf{89.4} & 82.2 & 81.7 & \textbf{81.1} \\
\bottomrule
\end{tabular}
\caption{Precision, recall and F1 scores for new information detection on the English-Chinese test set.}
\label{tab:results-divergence-detection-test-zh}
\end{table}

\section{Dataset Construction} \label{sec:appendix-dataset-construction}

\paragraph{Wikipedia paragraph selection}

Pywikibot\footnote{v. 8.0.1, https://pypi.org/project/pywikibot/} was used to download articles which had versions in English, Spanish, Chinese and Hindi.\footnote{Download date: March 22, 2023.} These were split into sections and paragraphs with wikitextparser.\footnote{v. 0.51.1, https://pypi.org/project/wikitextparser/} 

After selecting paragraph pairs, we further filtered the data according to length and paragraph similarity score. %
We only kept those with English paragraphs containing between 86 and 1000 characters, (Spanish, Hindi) paragraphs containing between 120 and 1000 characters, and a similarity score between .63 and .95. For Chinese-English paragraphs, we removed pairs where the Chinese paragraphs had over 250 characters.

\paragraph{Annotation Process} 

We recruited workers from Upwork, selecting those who were either bilingual or fluent in either language, and who had translation experience between the languages of interest. To ensure quality control, workers had to pass a qualification round consisting of 14 paragraph pairs (i.e., 7 pairs, but annotating both directions). These qualification rounds also served to give feedback to the annotators. Four annotators were chosen for English-Spanish and English-Chinese, and three annotators were chosen for English-Hindi. Annotators were paid \$300 for every 140 paragraph pairs (70 paragraph pairs, in both directions), at an estimated hourly rate of \$10-\$25. Annotators were hired from Argentina, Colombia, India, China and the US. All annotators had at least an undergraduate degree, and seven had post-graduate degrees.

Annotators were presented with each (Spanish, Hindi, Chinese) paragraph first and asked to annotate the related English paragraph; then the order of the paragraphs were flipped and they were asked to annotate the (Spanish, Hindi, Chinese) paragraph. Annotators were able to reject (toss out) paragraphs that were too dissimilar, for cases where the entire paragraph was new (in each direction) or when the paragraphs had superficial similarities but were about completely different subjects. They were also given the option to leave a comment for each paragraph pair in order to leave feedback or outline their thought process. The instructions given to annotators are in Appendix \ref{sec:appendix-dataset-instruction}. Prodigy \citep{prodigy_montani_honnibal} was used for the annotation interface, shown in Appendix \ref{sec:appendix-dataset-interface}.

\section{Implementation Details} \label{sec:appendix-implementation-details}

\paragraph{Alignment}  We use SimAlign \citep{jalili-sabet-etal-2020-simalign}, an MT aligner based on comparing cosine similarities of mBERT embeddings. SimAlign was chosen because its performance is comparable to the best supervised aligners such as fastalign/IBM2 \citep{dyer-etal-2013-simple}, efmaral/eflomal \citep{ostling2016efficient} and Giza++/IBM4 \citep{och-ney-2003-systematic}.\footnote{Giza++, fastalign and efmaral refer to different implementations of the original systems.} We use the \emph{argmax} method of SimAlign, and tune the null threshold $\tau$ on our dev set in order to maximize F1. For \mbox{\textsc{es-en}} and \mbox{\textsc{en-es}} we used $\tau=0.9997$, for \mbox{\textsc{hi-en}} and \mbox{\textsc{en-hi}} we used $\tau=0.99979$, and for \mbox{\textsc{zh-en}} and \mbox{\textsc{en-zh}} we used $\tau=0.99976$.   Evaluations with Alignment were done on a laptop with 32 GB of RAM and no GPUs.

\paragraph{SLR-NLI}  We retrained the SLR-NLI BERT-based model.\footnote{Without e-SNLI supervision, using the defaults in \url{https://github.com/joestacey/snli_logic}} 
We sentence-segment the target paragraph and run SLR-NLI on each (paragraph, sentence) pair. When the target paragraph is non-English, any predicted spans are mapped back to the source paragraph using an MT aligner (SimAlign with \emph{itermax}). 
We used SLR-NLI with combinations of 2 consecutive spans.\footnote{See the discussion in Section~2.2 of \citet{stacey2022logical}.} The threshold for selecting neutral and contradiction spans was tuned on the development set. We used thresholds of $0.15$ for \mbox{\textsc{es-en}} and \mbox{\textsc{en-es}}, $0.20$ for \mbox{\textsc{hi-en}}, $0.10$ for \mbox{\textsc{en-hi}}, $0.15$ for \mbox{\textsc{zh-en}}, and $0.25$ for \mbox{\textsc{en-zh}}.  Evaluations with SLR-NLI were performed on a laptop with 32 GB of RAM and no GPUs.

\paragraph{NLI Attribution}  We used a BERT-based \citep{devlin-etal-2019-bert} NLI model trained on MNLI\footnote{\url{https://huggingface.co/gchhablani/bert-base-cased-finetuned-mnli}} \citep{williams-etal-2018-broad}. For our attribution method, we use integrated gradients \citep{sundararajan-et-al-2017}. We chose this model and attribution method after preliminary experiments comparing three different attribution methods (Saliency, InputXGradients and Integrated Gradients) and two different models (BERT and deBERTa\footnote{Specifically deBERTa trained on a mix of NLI datasets: \url{https://huggingface.co/MoritzLaurer/DeBERTa-v3-base-mnli-fever-docnli-ling-2c}}) on the dev set of \mbox{\textsc{es-en}}.

NLI Attribution experiments were done on one NVidia Titan RTX GPU. Thresholds for selecting tokens based on their attribution scores were tuned on the development set. We used thresholds of $0.03052$ for \mbox{\textsc{es-en}}, $0.02263$ for \mbox{\textsc{en-es}}, $ .02260$ for \mbox{\textsc{hi-en}} and \mbox{\textsc{en-hi}}, $0.02260$ for \mbox{\textsc{zh-en}}, and $0.02470$ for \mbox{\textsc{en-zh}}. 

Intuitively, spans which contain new information not present in the source paragraph should cause NLI models to classify the hypothesis as \emph{neutral} or \emph{contradiction}. SLR-NLI is designed explicitly to find these spans, while attribution methods may surface tokens which are neutral with higher attribution scores. Since both SLR-NLI and the token attribution model are monolingual (English) models, for both methods we first translate either the source (for \mbox{\textsc{*-en}} pairs) or the target (for \mbox{\textsc{en-*}} pairs) paragraph to English using Google Translate.\footnote{Future work can consider inherently cross-language NLI models.} When the language of the source paragraph is non-English, we use its translation as the premise, and when the target language is non-English, we translate the target paragraph to English to use as the hypothesis\footnote{Or, more precisely, each sentence of the translated target paragraph is a hypothesis; we run each method over all (paragraph, sentence) pairs and aggregate the results.} for the NLI model, and any localized spans must be mapped via MT alignment back to the tokens of the target paragraph. For NLI Attribution, we follow \citet{zaman2022multilingual} and do this by summing the attribution scores for all translation tokens which map onto a target paragraph token. %

\paragraph{LLMs} When testing \textbf{GPT-3.5-turbo} and \textbf{GPT-4}, we specifically used gpt-3.5-turbo-0613 and gpt-4-0613, since these models will not be updated.\footnote{\url{https://platform.openai.com/docs/models/continuous-model-upgrades}, last accessed Oct. 15, 2023.}. We used the 7B version of \textbf{Llama-2-chat} \citep{touvron2023llama}, the 7.1B version of \textbf{BLOOMZ} \citep{muennighoff-etal-2023-crosslingual} and the 7.5B version of \textbf{XGLM} \citep{lin-etal-2022-shot}. BLOOMZ is an instruction-tuned model, while XGLM is a non-instruction tuned autoregressive LM. We used prompts that specify the annotation task in depth, given in Appendix \ref{sec:appendix-gpt-prompt}.

We obtained similar performance for the GPT models whether we presented the data as (paragraph, paragraph) pairs or (paragraph, sentence) pairs, so we only report the paragraph-level version. However, sentence segmenting and running the LLMs for (paragraph, sentence) pairs was slightly more effective for the smaller LMs, so we report the sentence-level version for these. For GPT-3.5-turbo and GPT-4 we used a temperature of 0.7 and top-p of 1, while for the smaller language models we used greedy decoding.

For the BLOOMZ, XGLM and Llama-2-chat experiments, we used a NVIDIA RTX A6000 GPU. Llama-2-chat took under 30 minutes per experiment, while XGLM and BLOOMZ took under 1 hour per experiment.

\onecolumn
\section{Dataset Examples} \label{sec:appendix-paragraph-examples}

\begin{figure}[h!]
\centering
    \includegraphics[width=\linewidth, trim=0 80 0 40]{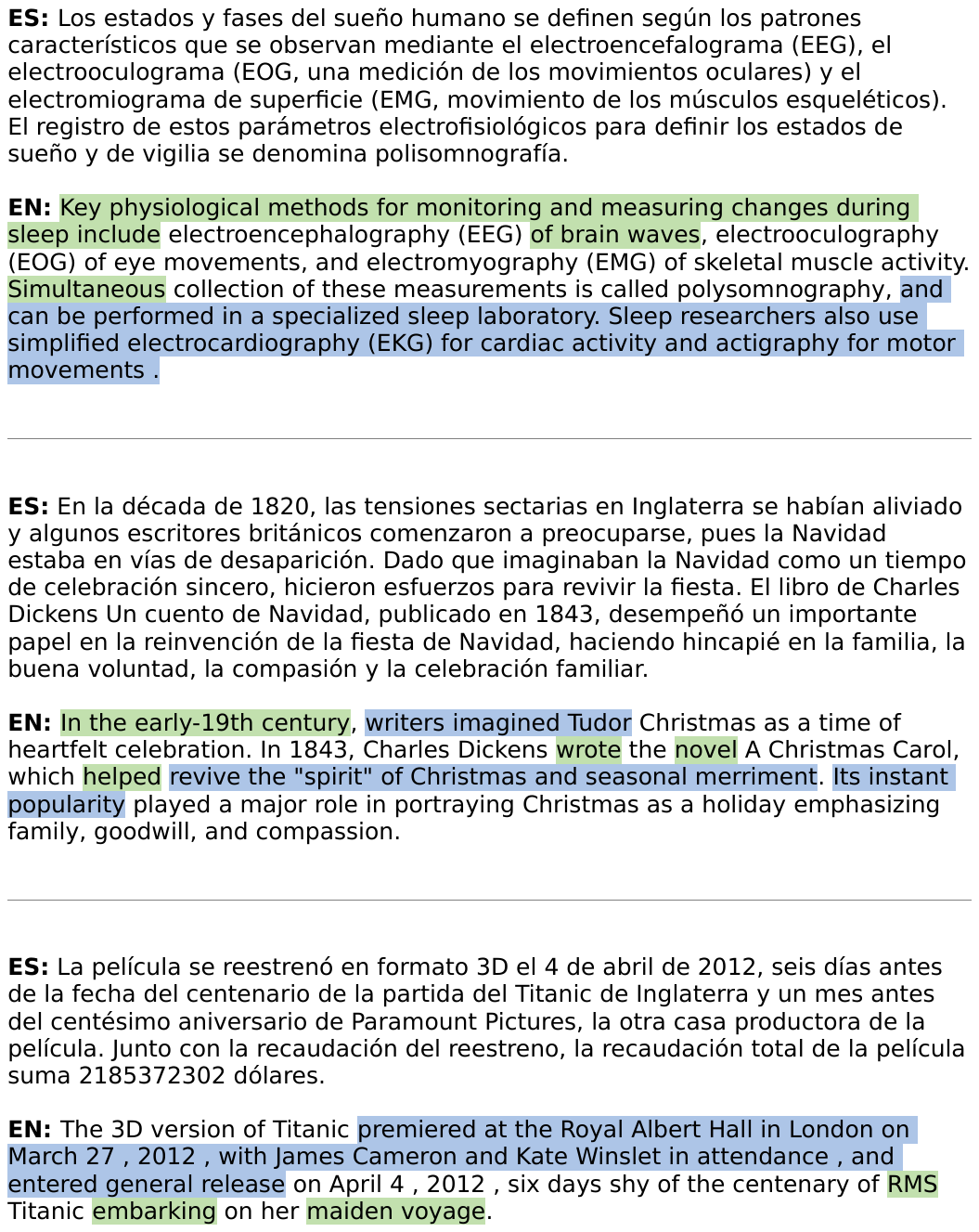}
    \caption{Three examples of paragraph pairs from the \mbox{\textsc{es-en}} portion of \datasetname~annotated with spans for \emph{new information} (\ns{blue}) and \emph{inferable} (\is{green}).}
    \label{figs:paragraph-examples}
\end{figure}

Figure~\ref{figs:paragraph-examples} shows three examples from our dataset.

\newpage
\section{Annotation Interface} \label{sec:appendix-dataset-interface}

\begin{figure}[h]
\centering
    \includegraphics[width=\linewidth]{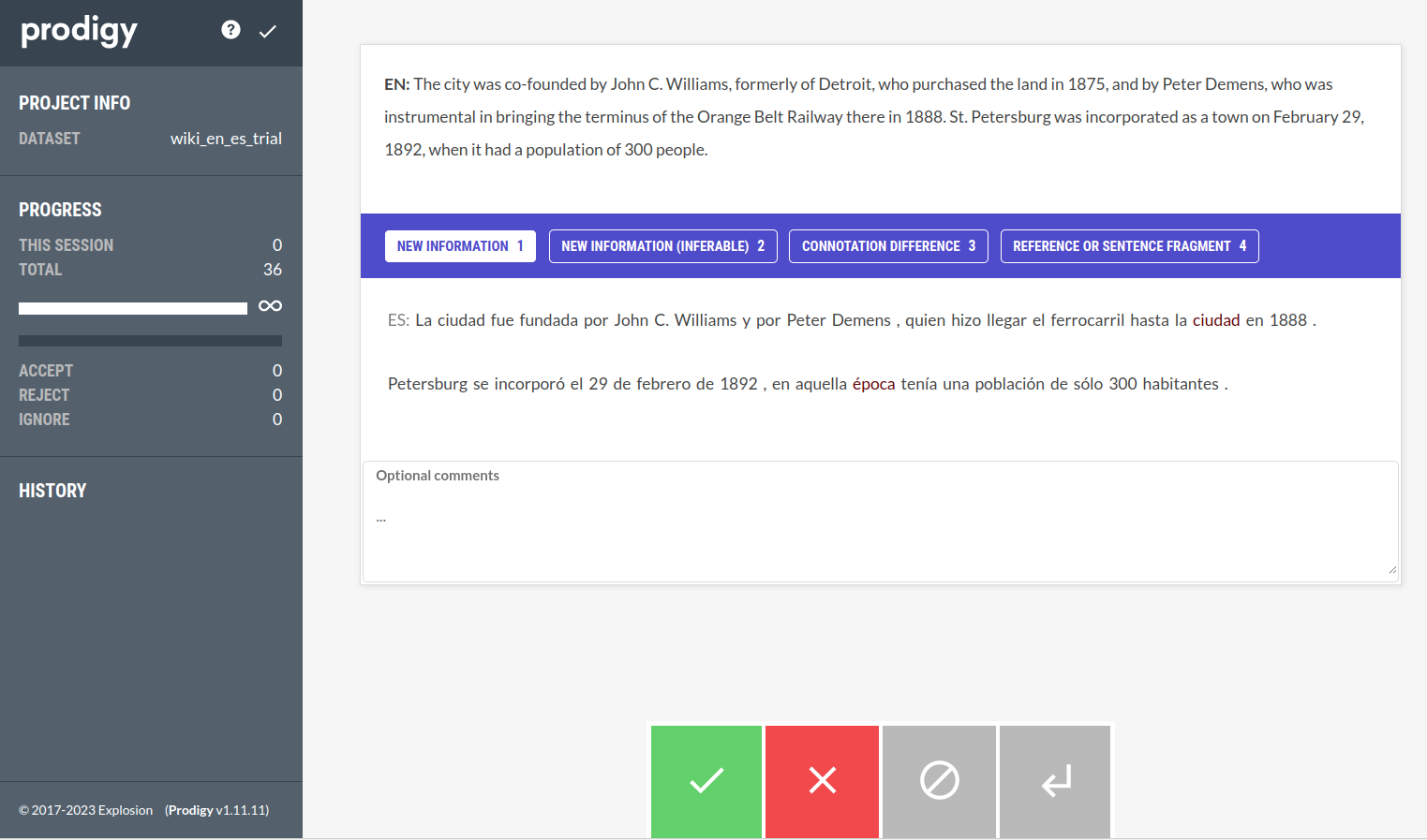}
    \caption{Screenshot of the interface used to annotate the dataset. We highlighted in red tokens based on the output of a word aligner in order to enable annotators to more easily spot differences. However, annotators were cautioned that these highlights were merely suggestions.}
    \label{figs:annotation-interface}
\end{figure}

\newpage
\section{Prompt for LLMs}\label{sec:appendix-gpt-prompt}

The prompts used for the LLMs in our experiments are shown below in Figures \ref{figs:prompt1} and \ref{figs:prompt2}:

\begin{figure}[h!]
\centering
    \includegraphics[width=\linewidth, trim=50 360 50 60]{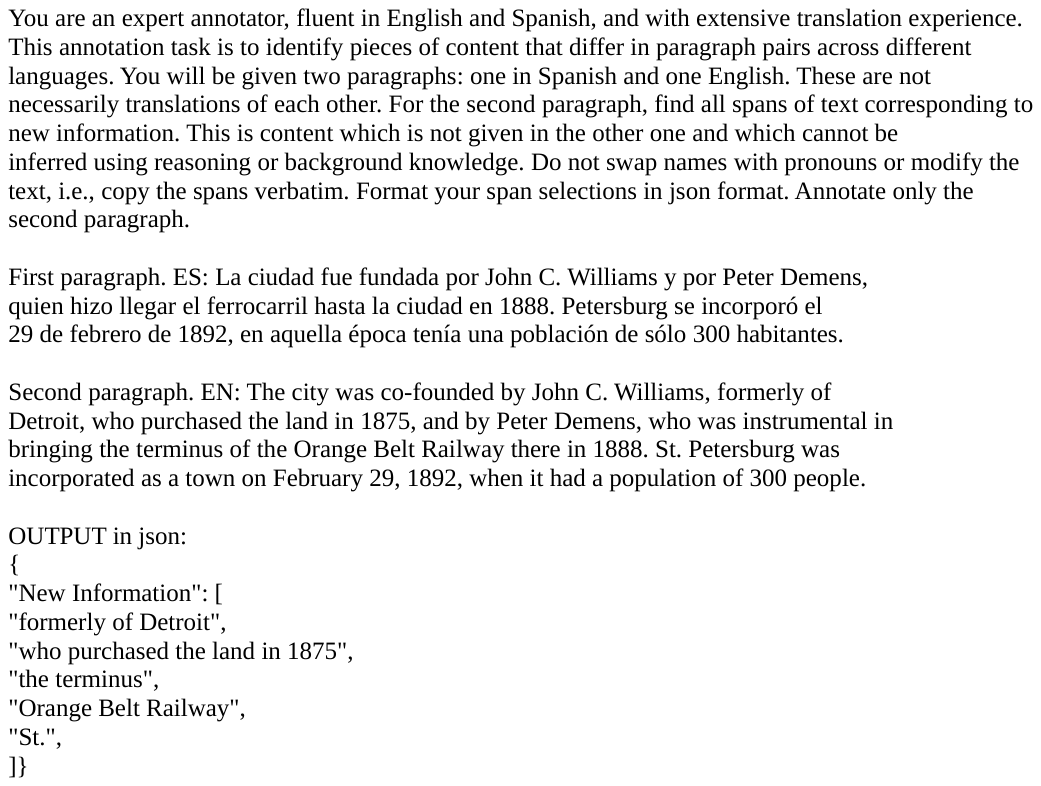}
    \caption{One-shot prompt used for the evaluating GPT-3.5-turbo and GPT-4 on the \mbox{\textsc{es-en}} portion of the dataset. For the other language directions we translated the output spans and source and target paragraphs appropriately. For the smaller LMs, we had it output a list rather than a json, since they struggled in producing valid json.}
    \label{figs:prompt1}
\end{figure}

\newpage

\begin{figure}[h!]
\centering
    \includegraphics[width=\linewidth, trim=50 210 50 60]{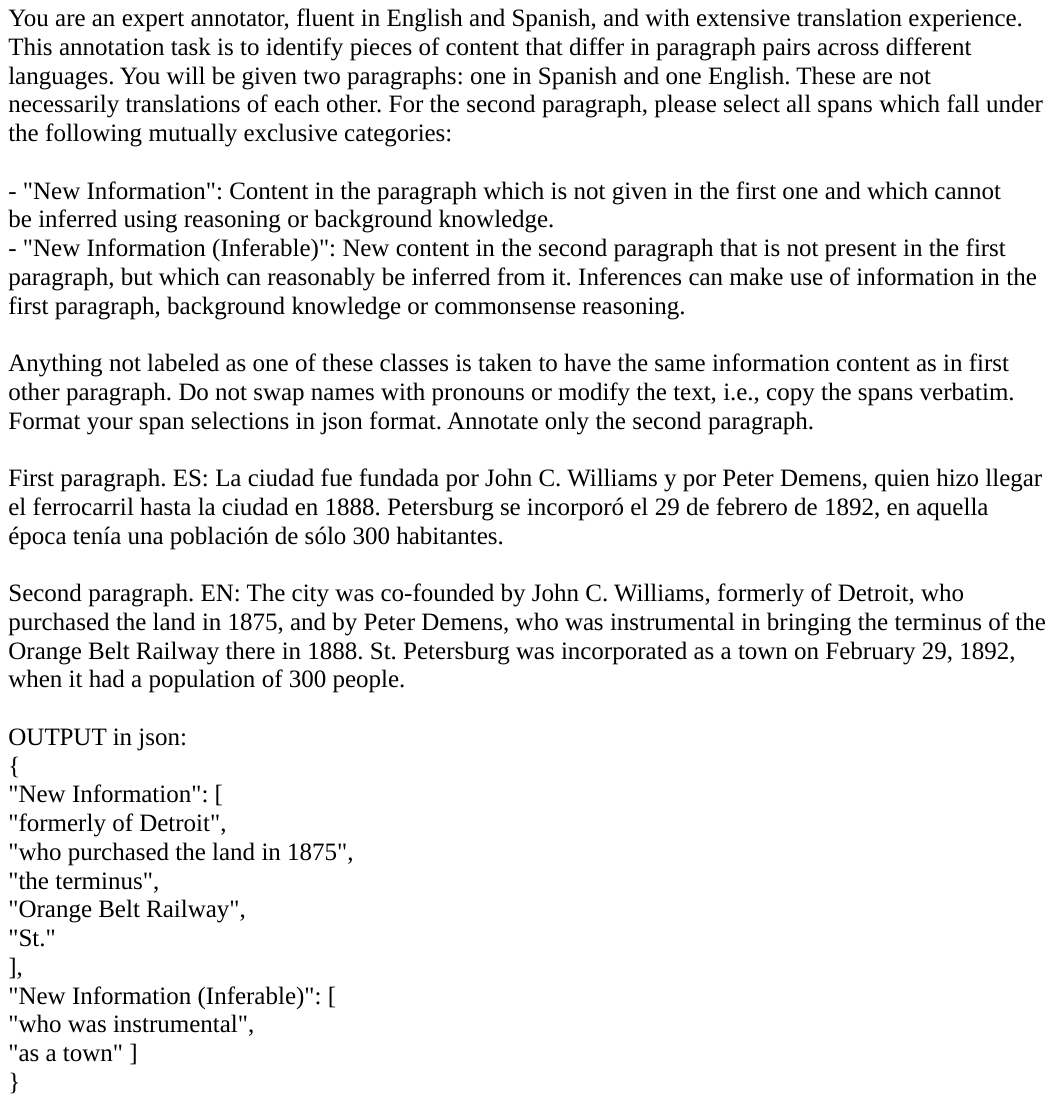}
    \caption{One-shot prompt used for evaluating GPT-4 on three-way (new, same, inferable) classification task on the \mbox{\textsc{es-en}} portion of the dataset. For the other language directions we translated the output spans and source and target paragraphs appropriately.}
    \label{figs:prompt2}
\end{figure}

\section{Annotator Instructions} \label{sec:appendix-dataset-instruction}

\includepdf[pages=-, scale=0.90]{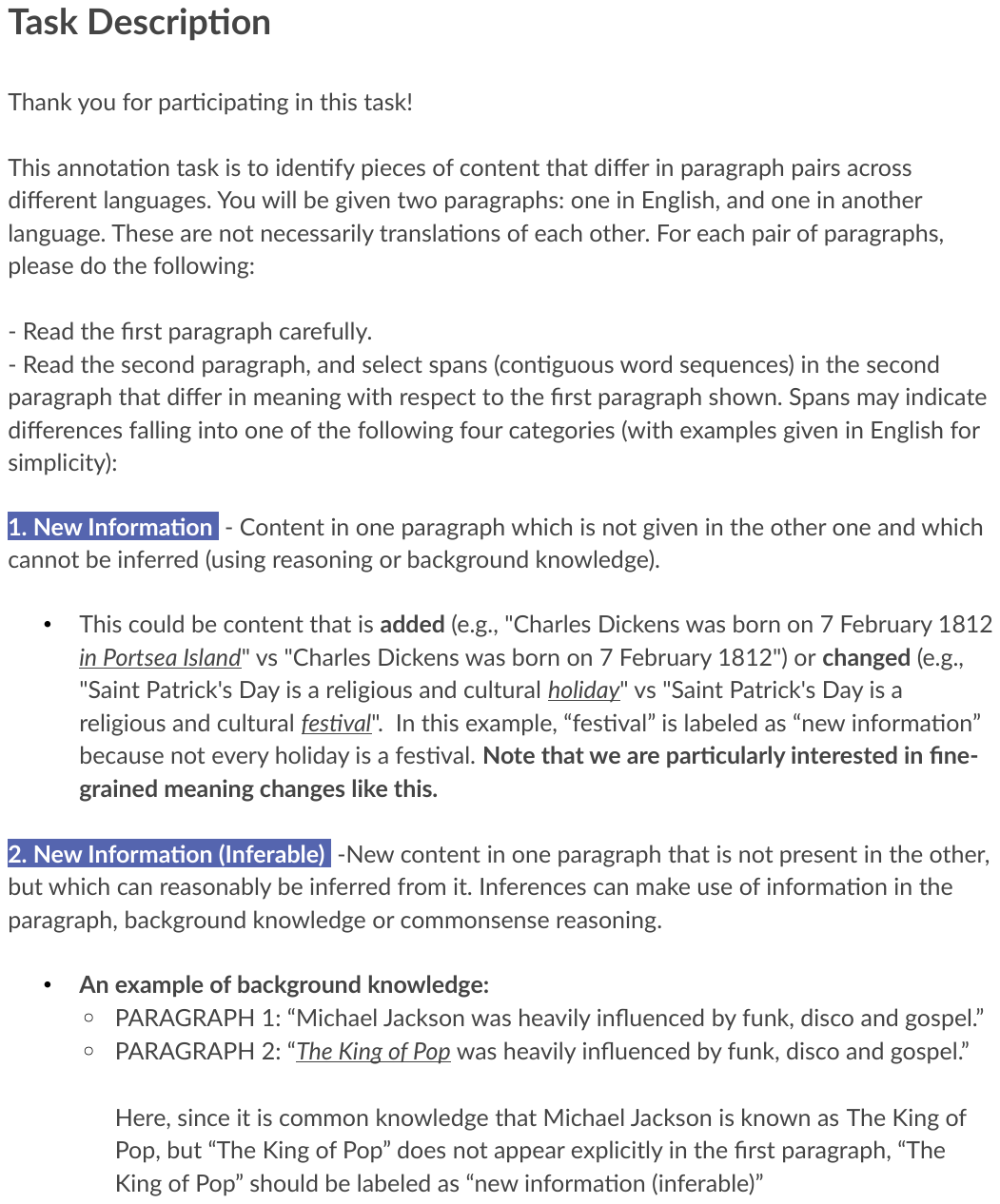}

\section{Examples of Inferable Span Disagreement} \label{sec:appendix-inferable-span-disagreement}

\begin{table}[h!]
    \centering
    \scriptsize
\begin{tabularx}{\textwidth}{Xp{3.5cm}l}
    \toprule
    \textbf{Source paragraph (truncated) } & \textbf{Target paragraph (truncated)} & \textbf{Inferable span?} \\
    \midrule
     Las liebres son solitarias, aunque no les importa en absoluto la presencia de otras liebres en los alrededores. \isr{Tan solo se producen peleas durante la época de celo} (variable según especies), que pueden llegar a ser hasta cierto punto cómicas en algunas especies. \isr{Las liebres europeas de sexo masculino apenas comen durante este período (primavera)}, y pasan el día luchando con sus rivales...  %
     & Normally a \is{shy} animal, the European brown hare \is{changes its behavior in spring}, when it can be seen in daytime chasing other hares...
     & \ns{No}, Yes \\
    \midrule
    Tercera edad o senectud es un término antroposocial que hace referencia a \isr{las últimas décadas de la vida}, en la que uno \isr{se aproxima a la edad máxima que el ser humano puede vivir}.
    & \is{Old age} is \is{the range of ages nearing} and surpassing \is{the life expectancy} of human beings; \is{it is the end of the human life cycle}...
    & \samespan{No}, Yes, \samespan{No}, Yes \\
    \midrule
    \isr{El Templo de Júpiter del Capitolino fue comenzado por Tarquinio Prisco y completado por el último rey de Roma}, Tarquinio el Soberbio, aunque fue inaugurado, según una tradición registrada por los historiadores, el 13 de septiembre, al comienzo de la época republicana...
    & The building was supposedly begun by \is{king} Tarquinius Priscus, completed by the last king (Tarquinius Superbus) and inaugurated.
    & Yes \\
    \midrule
    La Semana Santa, y la Pascua en particular, está ligada a través de la última cena y la crucifixión de Jesús a la Pésaj (Pascua Judía) y \isr{al Éxodo del pueblo hebreo narrado en el Antiguo Testamento}...
    & Easter is linked to Passover and the Exodus \is{from Egypt recorded} in the Old Testament through the Last Supper, sufferings, and crucifixion of Jesus that preceded the resurrection.
    & Yes \\
    \midrule
    Luego de ello, Affleck \isr{comenzó a salir con} Jennifer Lopez en julio \textbf{de 2002 tras protagonizar juntos Gigli} (2003). También trabajaron en Jersey Girl (2004) y en el videoclip de «Jenny from the Block». Su relación, la cual fue apodada «Bennifer» y considerada como una superpareja, atrajo una atención masiva por parte de los medios y se generaron una gran cantidad de rumores sobre aspectos personales de ambos. La pareja se comprometió en noviembre de 2002 y tenían una boda prevista para el 14 de septiembre de 2003, pero fue pospuesta apenas cuatro días antes del evento a causa del acoso de los paparazzi. Finalmente \isr{se separaron en enero de 2004} en buenos términos.
    &
    Affleck first dated Jennifer Lopez from 2002 \is{to 2004}. They became friends \is{on the set} of Gigli in December 2001, having previously encountered each other at industry parties. \is{They began a romantic relationship} in July 2002 when Lopez filed for divorce from her second husband, Cris Judd.
    &
    Yes, Yes, Yes \\
    \midrule
    La pelvis es la región anatómica inferior del tronco. Siendo una cavidad, la pelvis es un embudo osteomuscular que se estrecha hacia abajo, \isr{limitado por} el hueso sacro, el cóccix y los coxales (que forman la cintura pélvica) y \isr{los músculos de la pared abdominal inferior y del perineo}. Limita un espacio \isr{llamado cavidad pélvica, en donde se encuentran órganos importantes}...
    &
    The pelvic region of the trunk is the lower part of the trunk, \is{between the abdomen and the thighs}. \is{It includes several structures}: the bony pelvis, the pelvic cavity, the pelvic floor, and the perineum.
    &
    Yes, Yes \\
    \midrule
    Ese mismo año conoció a uno de los grandes amores de su vida, Charles-Joseph Lamoral, \isr{príncipe} de Ligne.
    &
    Early in Bernhardt 's career, she had an affair with a Belgian \is{nobleman}, Charles-Joseph Eugène Henri Georges Lamoral de Ligne (1837–1914), son of Eugène, 8th Prince of Ligne, with whom she bore her only child, Maurice Bernhardt (1864–1928).
    &
    Yes \\
    \midrule
    El 19 de abril, Malcolm X concluyó el Hajj, dando las siete vueltas alrededor de la Kaaba, bebiendo del Pozo de Zamzam y corriendo siete veces a través de las colinas de Al-Safa y Al-Marwah. Según su autobiografía, este viaje le permitió ver a los musulmanes \isr{de diferentes razas} que interaccionan como iguales \isr{y llegó a creer} que el islam puede superar los problemas raciales.
    &
    MalcolmX \is{later said} that seeing Muslims of \is{"all colors, from blue-eyed blonds to Black-skinned Africans}," interacting as equals \is{led him to see} Islam as a means by which racial problems could be overcome.
    &
    Yes, \ns{No}, No \\
    \midrule
    Las ovejas han tenido una fuerte presencia en la cultura de muchos países, especialmente en las zonas donde constituyen el tipo más común de ganado. En la literatura, especialmente en las fábulas, son las representantes típicas de la bondad, \isr{mansedumbre} y las pocas luces, en contraposición con el lobo o el zorro...
    &
    In the English language, to call someone a sheep or ovine may allude that they are timid and \is{easily led}.
    &
    Yes \\
    \midrule
    Después de \isr{la conquista} del Imperio aqueménida, a manos de Alejandro Magno, y más tarde, tras la caída de los partos, el Imperio sasánida gobernó \isr{el norte y el sur} del golfo, manteniendo la Ruta de la Seda.
    &
    Following the \is{fall} of Achaemenid Empire, and after the fall of the Parthian Empire, the Sassanid Empire ruled the northern \is{half} and at times the southern \is{half} of the \is{Persian} Gulf... 
    &
    Yes, \ns{No}, \ns{No}, Maybe \\
    \midrule
    El kriya yoga es \isr{la forma práctica} de las doctrinas del yoga, la unión \isr{con Dios} mediante la devoción activa y la realización correcta de los deberes diarios.
    &
    The "science" of Kriya Yoga is the foundation of Yogananda's teachings. \is{An ancient} \is{spiritual practice}, Kriya Yoga is union \is{(yoga)} with the Infinite.
    & Maybe, Yes, \samespan{No} \\
    \bottomrule
\end{tabularx}
    \caption{Examples of spans labeled as inferable (\is{green}) in the \mbox{\textsc{es-en}} portion of \datasetname where not all annotators agreed on the span label. The right column shows, for each span, whether we judge the span to be inferable (\emph{Yes}), not inferable (\emph{No}, shown in \ns{blue} in cases where \emph{new} is a more appropriate label), and \emph{Maybe} for cases where our the answer depends on how much domain-specific background knowledge one draws from to make the inference. The most relevant parts of the Spanish paragraph for each judgement are shown in \isr{bold}.}
    \label{tab:text_pairs_inf_labels}
\end{table}

\clearpage

\begin{table}[!t]
    \centering
    \scriptsize
\begin{tabularx}{\textwidth}{Xp{3.5cm}l}
    \toprule
    \textbf{Source paragraph (truncated) } & \textbf{Target paragraph (truncated)} & \textbf{Inferable span?} \\
    \midrule
    Rachel Louise Carson (27 de mayo de 1907 - 14 de abril de 1964) fue una bióloga marina y conservacionista estadounidense que, a través de \isr{la publicación de} Primavera silenciosa en 1962 y otros escritos, \isr{contribuyó a la puesta en marcha de la moderna conciencia ambiental}.
    &
    Rachel Louise Carson (May 27, 1907 – April 14, 1964) was an American marine biologist, \is{writer}, and conservationist whose \is{influential} book Silent Spring (1962) and other writings are \is{credited} with \is{advancing} the global environmental movement. 
    & Yes, Yes, Yes, Yes \\
    \midrule
    La búsqueda de \isr{los rasgos} de líderes han sido una constante en todas las culturas durante siglos. Escrituras filosóficas como la República de Platón o las Vidas de Plutarco han explorado una pregunta básica: «¿Qué cualidades distinguen a un líder?».
    &
    The search for the \is{characteristics} or traits of leaders has continued for centuries. Philosophical writings from Plato's does not use the word "leadership"...
    &
    No \\
    \midrule
    En \isr{la década de 1820}, las tensiones sectarias en Inglaterra se habían aliviado y algunos escritores británicos comenzaron a preocuparse, pues la Navidad estaba en vías de desaparición. Dado que imaginaban la Navidad como un tiempo de celebración sincero, hicieron esfuerzos para revivir la fiesta. \isr{El libro de Charles Dickens Un cuento de Navidad, publicado en 1843}, \isr{desempeñó un importante papel} en la reinvención de la fiesta de Navidad, haciendo hincapié en la familia, la buena voluntad, la compasión y la celebración familiar.
    &
    \is{In the early-19th century}, writers imagined Tudor Christmas as a time of heartfelt celebration. In 1843, Charles Dickens \is{wrote} the \is{novel} A Christmas Carol, which \is{helped} revive the "spirit" of Christmas and seasonal merriment.
    &
    Yes, Yes, Yes, Yes \\
    \midrule
    La longitud es una medida de una dimensión (lineal; por ejemplo la distancia en m), mientras que el área es una medida de dos dimensiones (al cuadrado; por ejemplo m²), y el volumen es una medida de tres dimensiones (cúbica; por ejemplo m³).
    &
    Length is the measure of one \is{spatial} dimension, whereas area is a measure of two dimensions (\is{length} squared) and volume is a measure of three dimensions (\is{length} cubed).
    &
    Yes, Yes, Yes \\
    \midrule
    Es \isr{venerado} como santo por la Iglesia evangélica luterana en Estados Unidos (Calendario de Santos Luterano) \isr{y la Iglesia anglicana}. Su festividad se conmemora el 31 de marzo.
    &
    Donne is \is{remembered} in the Calendar of Saints of the \is{Church of England}, the Episcopal Church liturgical calendar and the Calendar of Saints of the Evangelical Lutheran Church in America for his life as both poet and priest.
    & Yes, No \\
    \midrule
    El ácido láctico, o su forma ionizada, el lactato (del lat. lac, lactis, leche), también conocido por su nomenclatura oficial ácido 2-hidroxi-propanoico o ácido $\alpha$-hidroxi-propanoico, es un compuesto químico que desempeña importantes roles en varios procesos bioquímicos, como la fermentación láctica. Es un ácido carboxílico, \isr{con un grupo hidroxilo en el carbono adyacente al grupo carboxilo}, lo que lo convierte en un ácido $\alpha$-hidroxílico (AHA) de fórmula H3C-CH(OH)-COOH (). \isr{En solución puede perder el hidrógeno unido al grupo carboxilo y convertirse en el anión lactato.}
    &
    Production includes both artificial synthesis as well as natural sources. Lactic acid is an alpha-hydroxy acid (AHA) due to the \is{presence} of a hydroxyl group adjacent to the carboxyl group. It is used as a synthetic intermediate in many organic synthesis industries and in various biochemical industries. \is{The conjugate base} of lactic acid is called lactate (or the lactate anion).%
    &
    No, Maybe \\  
    \bottomrule
\end{tabularx}
    \caption{Examples of spans labeled as inferable (\is{green}) in the \mbox{\textsc{es-en}} portion of \datasetname where not all annotators agreed on the span label. The right column shows, for each span, whether we judge the span to be inferable (\emph{Yes}), not inferable (\emph{No}, shown in \ns{blue} in cases where \emph{new} is a more appropriate label), and \emph{Maybe} for cases where our the answer depends on how much domain-specific background knowledge one draws from to make the inference. The most relevant parts of the Spanish paragraph for each judgement are shown in \isr{bold}.}
    \label{tab:text_pairs_inf_labels2}
\end{table}

\twocolumn

\end{document}